\documentclass[
  sora,
  strictfonts,
  numberedsections,
  onecolumn
]{mirrosarticle}

\usepackage{xspace}
\usepackage{url}
\usepackage{float} %
\usepackage{amsthm}
\newtheorem{assumption}{Assumption}
\newtheorem{property}{Property}
\colorlet{takeawayaccent}{MirrosAccent}
\definecolor{takeawaybg}{HTML}{F6F5F9}
\newtcolorbox{takeaway}{
  enhanced,
  breakable,
  sharp corners,
  frame hidden,
  boxrule=0pt,
  colback=takeawaybg,
  borderline west={2pt}{0pt}{takeawayaccent},
  boxsep=0pt,
  left=10pt,
  right=10pt,
  top=8pt,
  bottom=8pt,
  before skip=10pt,
  after skip=10pt,
  before upper={\textcolor{takeawayaccent}{\textbf{Takeaway:}}\enspace},
}

\setlist[itemize]{leftmargin=*}
\setlist[enumerate]{leftmargin=*}
\title{\texorpdfstring{S-Space: Exploring Spatial Workspace in Multimodal Language Models}{S-Space: Exploring Spatial Workspace in Multimodal Language Models}}
\runningtitle{S-Space: Exploring Spatial Workspace in Multimodal Language Models}
\author{\textbf{MirroS Technical Report}}
\pdfauthor{Zhiyuan Shi, Xiang Li, Xingzhuo Guo, Shangchen Miao, Weiliang Chen, Qiyu Dai, Jinshan Ren, Jiangran Lyu, Fangfu Liu, Mingsheng Long, Jialong Wu}
\reportid{MirroS Technical Report}
\date{October 8, 2026}
\metadata[Project Page]{\url{https://mirros.ai/blog/s-space}}
\metadata[Code]{\url{https://github.com/mirros-lab/s-space}}

\begin{document}
\begin{abstract}
Multimodal language models (MLLMs) exhibit promising spatial reasoning capabilities, yet the internal mechanisms supporting them remain poorly understood. We uncover \textbf{S-Space}, a linear representation subspace in which object-token representations encode continuous coordinates along horizontal, vertical, and distance axes. Although we identify these axes by differentiating spatial-answer scores with respect to object-token activations, S-Space is intrinsic rather than induced by specific data or prompts, remaining readable under non-spatial questions. We show that S-Space functions as an internal spatial workspace of MLLMs. Perception encodes object locations into it, forming spatial representations that can be directly read out to support accurate spatial judgments. Internal reasoning transforms it for changing viewpoints. Verbal responses selectively access it, evidenced by causal interventions that alter spatial judgments while largely preserving non-spatial responses. Case studies further show that S-Space assigns locations to unseen entities and the observer, and connects physical directions with abstract linguistic meanings, suggesting both the capabilities and limitations of large-scale language training. Finally, we use S-Space as a diagnostic tool for spatial reasoning failures. Direction-word readouts reveal systematic coupling between cardinal and image-relative reference frames. More importantly, a perception-computation decomposition exposes a key bottleneck: MLLMs can form capable spatial representations, yet fall short in the spatial transformations required to reason reliably across viewpoint changes.
\end{abstract}

\maketitle
\section{Introduction}\label{section-introduction}\label{section-s-space-spatial-workspace}

Multimodal large language models (MLLMs)~\citep{seed2026seed2,team2026kimi,qwen2026qwen35,google2026gemini35flash} have rapidly advanced in spatial intelligence \citep{yang2025thinking, chen2024spatialvlm}, demonstrating utility in reconstruction~\citep{yin2026vision, wang2026code}, planning~\citep{team2025gemini}, navigation~\citep{zhang2024navid}, and manipulation~\citep{brohan2023rt, yuan2024robopoint} in the physical world. Despite this growing range of applications, robust spatial reasoning remains a fundamental challenge for MLLMs, with substantial performance gaps persisting on tasks that humans solve effortlessly~\citep{chen2026babyvision, yin2025spatial, zhang2026spinbench}. Interpreting these limitations requires looking beyond end-to-end performance: \textit{external} behaviors such as answers and actions alone reveal little about where failures arise, while the \textit{internal} representations and mechanisms remain poorly understood and underdiagnosed.

How is spatial information represented inside an MLLM, and how does it shape the model's responses? Research on human visual cognition offers a useful perspective: there is a \textit{mental workspace} in which internal representations are maintained and manipulated in the human brain~\citep{dehaene1998neuronal, schlegel2013network}. While existing works have explored spatial encodings and representational biases in MLLMs~\citep{kang2026linear,min2026whyfar}, a simple and unified explanation of how spatial information is encoded, transformed, and ultimately used to produce responses remains to be established. We ask whether MLLMs similarly possess an internal spatial workspace akin to that of the human brain. Uncovering such a mechanism could help explain spatial reasoning failures and point toward targeted improvements.

\begin{figure}[H]
  \centering
  \includegraphics[width=\linewidth]{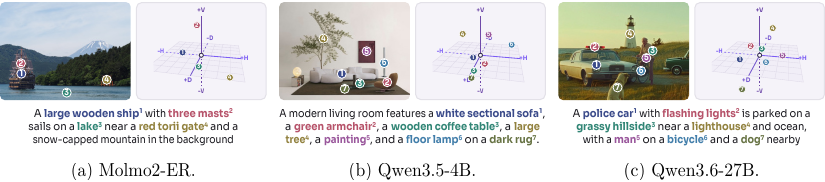}
  \caption{\textbf{Object coordinates read out from S-Space across MLLMs.} For each input image and caption, the intermediate representations of object-name tokens are linearly projected onto the three axes of S-Space to obtain continuous coordinates. Colors and numbers link object names, image regions, and their corresponding coordinates.}
  \label{fig:sspace-1}
\end{figure}

In this work, we uncover \textbf{S-Space}, a three-dimensional linear subspace of intermediate representations that functions as an internal spatial workspace in MLLMs. Using a simple gradient-based procedure, we identify fixed spatial orientations that generalize across datasets, tasks, and prompts, defining axes for left-right, above-below, and close-far directions. As illustrated in Figure~\ref{fig:sspace-1}, object locations can then be directly read out as continuous coordinates through linear projection of their representations~\citep{park2024linear}. Experiments demonstrate that S-Space accurately encodes object locations and provides spatial information that causally shapes the model's judgments, while remaining readable across prompts even when the model is answering questions unrelated to space.

With S-Space, we examine the spatial content represented within MLLMs. Through case studies, we show that the workspace brings visual evidence and linguistic meanings into a shared spatial geometry, extending its content beyond directly visible objects. Our final analysis comes as a diagnosis of spatial reasoning failures, building on S-Space's ability to separate spatial perception from subsequent computation. We study viewpoint-changing problems~\citep{zhang2026spinbench} by constructing an oracle that transforms S-Space coordinates, revealing that the encoded spatial information supports accurate answers even when the model fails to generate them. Direction-word readouts further reveal systematic coupling between cardinal and image-relative directions, offering insight into reference-frame confusion~\citep{yang2026mmsi}. These diagnoses expose a key bottleneck in spatial reasoning: MLLMs can form accurate perceptual representations yet fail to draw reliable spatial conclusions from them.

Our contributions can be summarized as follows:
\begin{itemize}
  \item To the best of our knowledge, we are the first to systematically identify and characterize an internal spatial workspace in MLLMs, revealing how spatial information can be encoded, updated, and causally used within the model's internal representations.
  \item We propose a simple gradient-based method for identifying S-Space, enabling accurate and generalizable continuous coordinate readouts along fixed spatial directions.
  \item We establish S-Space as a diagnostic tool for spatial reasoning failures, separating spatial perception from subsequent computation and exposing limitations in how models use available spatial information.
\end{itemize}

\section{Method}\label{section-method}

\begin{takeaway}
\textbf{S-Space} is a three-dimensional linear subspace of intermediate representations in MLLMs, in which object locations are linearly encoded and can be directly read out.
\end{takeaway}

In this section, we describe how we formulate, identify, and read out S-Space.

\begin{figure}[t]
  \centering
  \includegraphics[width=1.00\linewidth]{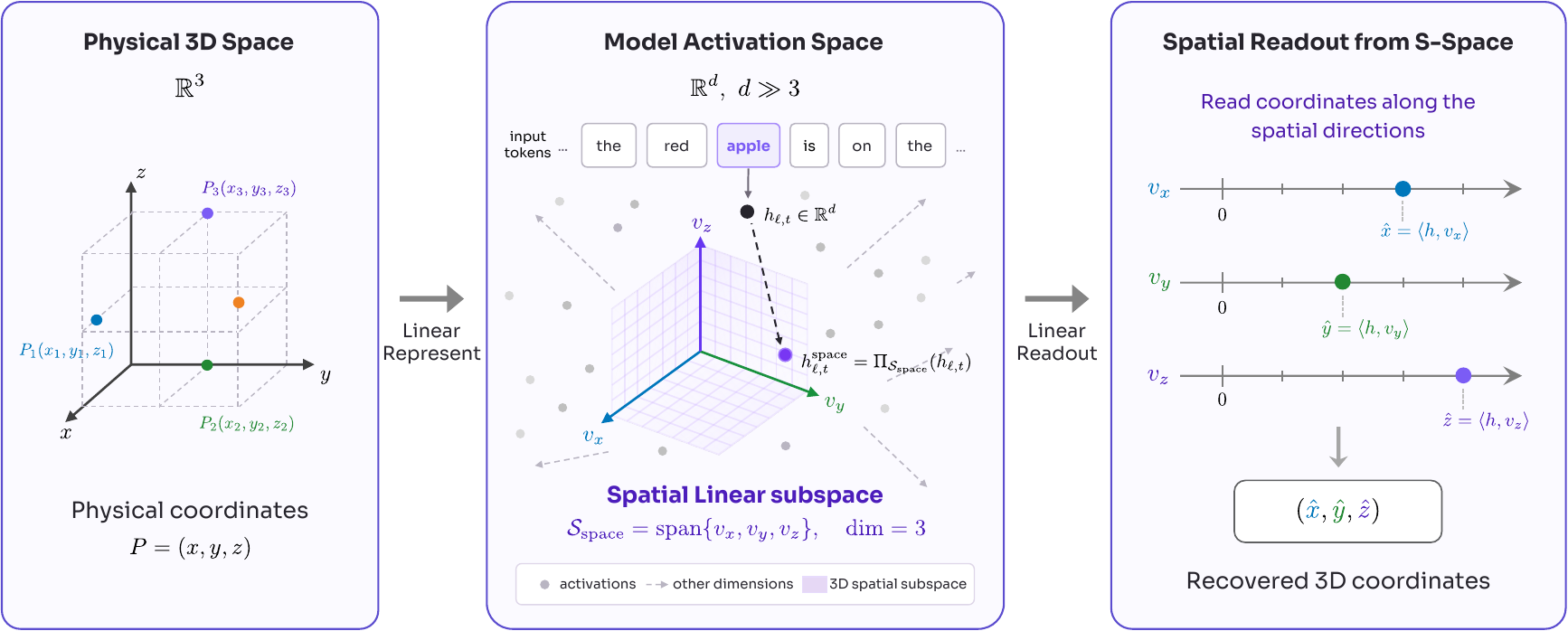}
  \caption{\textbf{Linear spatial representation in S-Space.} An object's spatial coordinates \textit{(left)} are linearly embedded in its high-dimensional token representation. Horizontal, vertical, and distance directions span a three-dimensional S-Space \textit{(middle)}. Projections onto them recover viewer-centered coordinates \textit{(right)}.}
  \label{fig:sspace-3}
\end{figure}

\subsection{S-Space Formulation}\label{sec:method-formulation}

Large language models are widely hypothesized to represent concepts linearly in their internal representations~\citep{elhage2022superposition, bricken2023monosemanticity}. Similarly, we hypothesize that an object's spatial coordinates, defined in a viewer-centered (or egocentric) reference frame\footnote{As we will see in Section~\ref{section-computation-scaling-transforming-s-space-through-reasoning} and \ref{app:image-order}, this reference frame is not fixed: it can be steered by context and progressively reoriented over the course of chain-of-thought reasoning.}, are linearly encoded in the hidden representation of its object-name token~\citep{kang2026linear}. 

Concretely, for a given model and layer $\ell$, we define S-Space as a three-dimensional linear subspace of hidden representations $h_{\ell, t} \in \mathbb{R}^{d_\mathrm{model}}$. S-Space is specific to the model and layer, but shared across all token positions $t$ within that layer. We search for S-Space in the model's middle layers, as early layers remain relatively anchored to the input representations, and later layers become increasingly aligned with next-token prediction for the production of a verbal response (see layer-wise analysis in Appendix~\ref{app:data}). We fix a layer $\ell$ and omit its index below for notational simplicity.

\subsection{Identifying S-Space}\label{sec:method-identification}

We identify three S-Space directions corresponding to the horizontal, vertical, and distance axes\footnote{For the horizontal and vertical axes, we use image-plane coordinates rather than world coordinates, without explicitly disentangling depth-dependent perspective effects. These projected coordinates suffice for most spatial reasoning questions considered in this work.} by differentiating the models' responses to relative spatial questions with respect to object-token activations~\citep{gurnee2026verbalizable}.

Specifically, each relative spatial question example $p$ consists of an image and a prompt comparing a \textit{query} object with a \textit{reference} along one axis \(g\in\{x,y,z\}\), e.g., ``Is the apple to the left or right of the cup?" Let $q$ and $r$ denote the respective token positions of the query and reference object names\footnote{When an object name consists of multiple tokens, we use its last token.}, we define an answer contrast for each axis as the corresponding logit difference:
\begingroup\small
\begin{equation}\label{eq:method-answer-contrasts}
s_{p,x}=\mathrm{logit}(\mathrm{right})-\mathrm{logit}(\mathrm{left}),\ 
s_{p,y}=\mathrm{logit}(\mathrm{above})-\mathrm{logit}(\mathrm{below}),\ 
s_{p,z}=\mathrm{logit}(\mathrm{close})-\mathrm{logit}(\mathrm{far}).
\end{equation}
\endgroup

The gradient $\nabla_{h_t}s_{p,g}$ therefore specifies the direction along which perturbing the object representation would most increase the corresponding answer contrast. We then form a role-antisymmetric gradient difference and average it across prompt templates, original/swapped object orders, and examples:
\begin{equation}
\label{eq:method-antisymmetric-jacobian}
a_{p,g} = \frac{\nabla_{h_{q}}s_{p,g}-\nabla_{h_{r}}s_{p,g}}{2},\quad 
v_g
=\operatorname{unit}\!\left(
\mathbb{E}_p\left[a_{p,g}\right]\right).
\end{equation}
Appendix~\ref{app:data} provides the prompt templates and implementation details.

The resulting unit vectors establish the three S-Space directions: $\mathcal S=\operatorname{span}\{v_x,v_y,v_z\}$.

\noindent\textbf{Theoretical sketch.} We offer two intuition for Equation~(\ref{eq:method-antisymmetric-jacobian}). (1) Relative spatial judgments treat the query and reference oppositely: a change that makes the query more to the right should favor \textit{right}, while the same change to the reference should favor \textit{left}. Taking the role-antisymmetric difference between gradients therefore reinforces such oppositely acting spatial signals. (2) Other effects arising from object identity, prompt phrasing, token roles, or scene context may not cancel for an individual example, so we further average across prompt templates, original/swapped object orders, and examples. The resulting direction captures the sensitivity that is most consistently associated with the queried spatial relation. See Appendix~\ref{section-theory} for detailed theoretical insights.

\subsection{Reading out S-Space}\label{sec:method-readout}

Once we identify the three axes, reading out S-Space requires only a forward pass. Importantly, this readout is not restricted to the spatial QA prompts used for identification: the same fixed directions apply to object-token representations elicited by other prompts and tasks (see Section~\ref{section-persistence}). We stack the directions into \(V\) and project each object token $t$ to get its coordinate in S-Space:
\begin{equation}\label{eq:method-coordinate-readout}
V=[v_x,v_y,v_z]\in\mathbb R^{d_{\mathrm{model}}\times 3},\qquad
c_t=V^\top h_t
\in\mathbb R^3.
\end{equation}

\noindent\textbf{Relative spatial judgment.} Reading out these coordinates supports spatial judgments without requiring the model to verbalize a response. For a relative spatial question, we compare the query and reference coordinates along the queried axis:
\begin{equation}\label{eq:method-relative-readout}
m_{g}=v_g^\top(h_q-h_r)=c_{q,g}-c_{r,g}.
\end{equation}
The sign of margin \(m_{g}\) directly selects the predicted endpoint: a positive value indicates right, above, or close, whereas a negative value indicates left, below, or far.

\section{S-Space as Internal Spatial Workspace}\label{section-the-spatial-world-in-s-space}

\begin{takeaway}
Across diverse contexts, S-Space functions as a \textbf{persistent} internal spatial workspace in MLLMs, where spatial information is accurately \textbf{encoded} into it (\S\ref{section-accuracy}), \textbf{transformed} through internal reasoning (\S\ref{section-computation-scaling-transforming-s-space-through-reasoning}), and causally \textbf{used} for verbal responses (\S\ref{section-causality}).
\end{takeaway}

In this section, we characterize S-Space as an internal spatial workspace in MLLMs by examining its accuracy, causal role, and persistence across prompting contexts. We identify S-Space for Molmo2-ER~\citep{fang2026molmoact2}, Qwen3.5-4B~\citep{qwen2026qwen35}, and Qwen3.6-27B~\citep{qwen2026qwen36} using constructed relative spatial questions from COCO~\citep{lin2014coco}. To assess the generalizability, we evaluate their readouts on CV-Bench~\citep{tong2024cambrian}, EmbSpatial~\citep{du2024embspatial}, and SpatialTunnel~\citep{min2026whyfar}. All experimental details are provided in Appendix~\ref{section-algorithm}.

\begin{figure}[t]
  \centering
  \includegraphics[width=\linewidth]{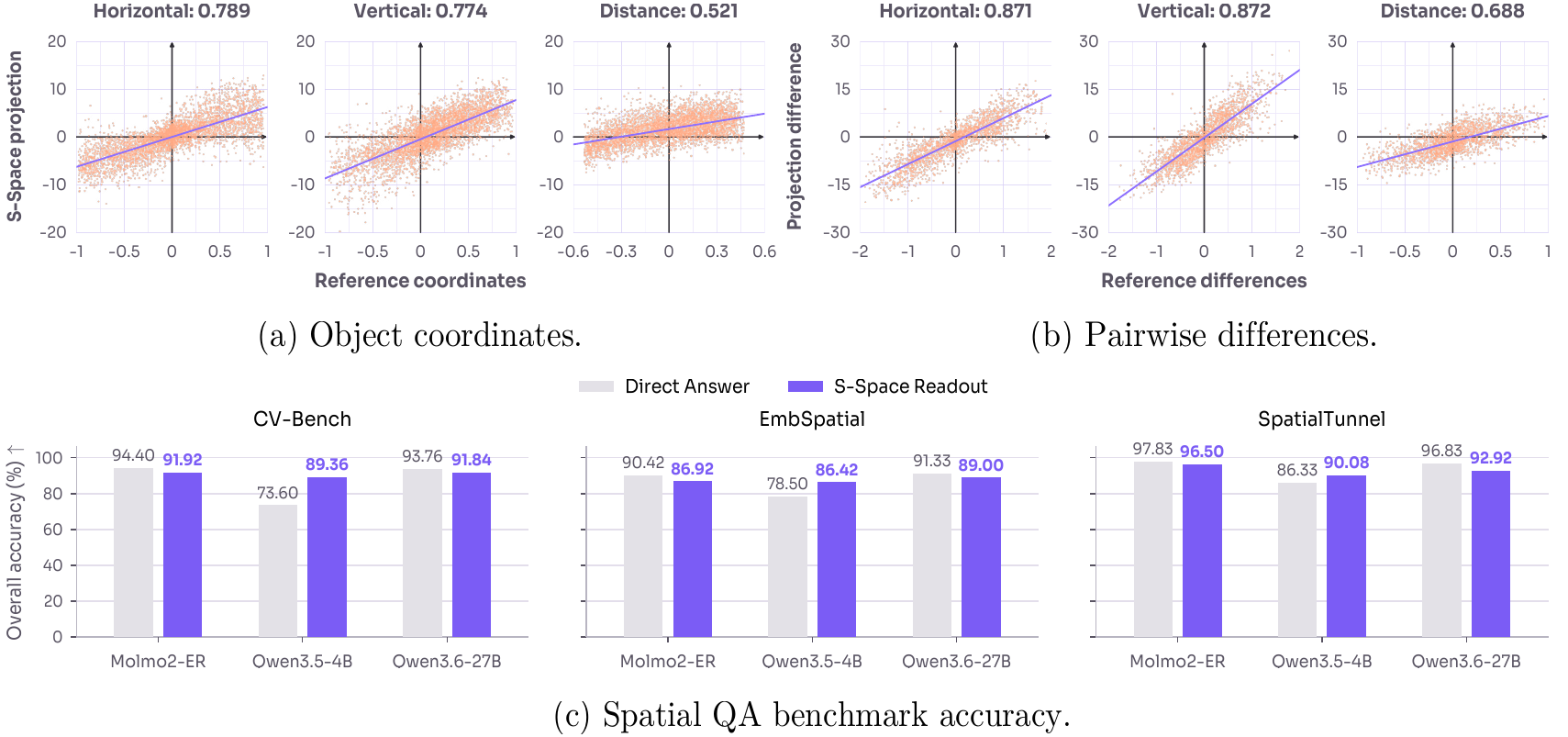}
  \vspace{-20pt}
  \caption{\textbf{Accuracy of S-Space.} \textit{(a-b)} S-Space coordinates read from layer 21 of Molmo2-ER versus ground-truth object positions, with Spearman correlations and least-squares fits. \textit{(c)} Spatial QA accuracy across benchmarks and models, comparing S-Space readouts with models' direct answers.}
  \label{fig:sspace-4}
  \label{fig:spatial-qa}
  \vspace{-6pt}
\end{figure}

\begin{figure}[t]
  \centering
  \includegraphics[width=1.0\linewidth]{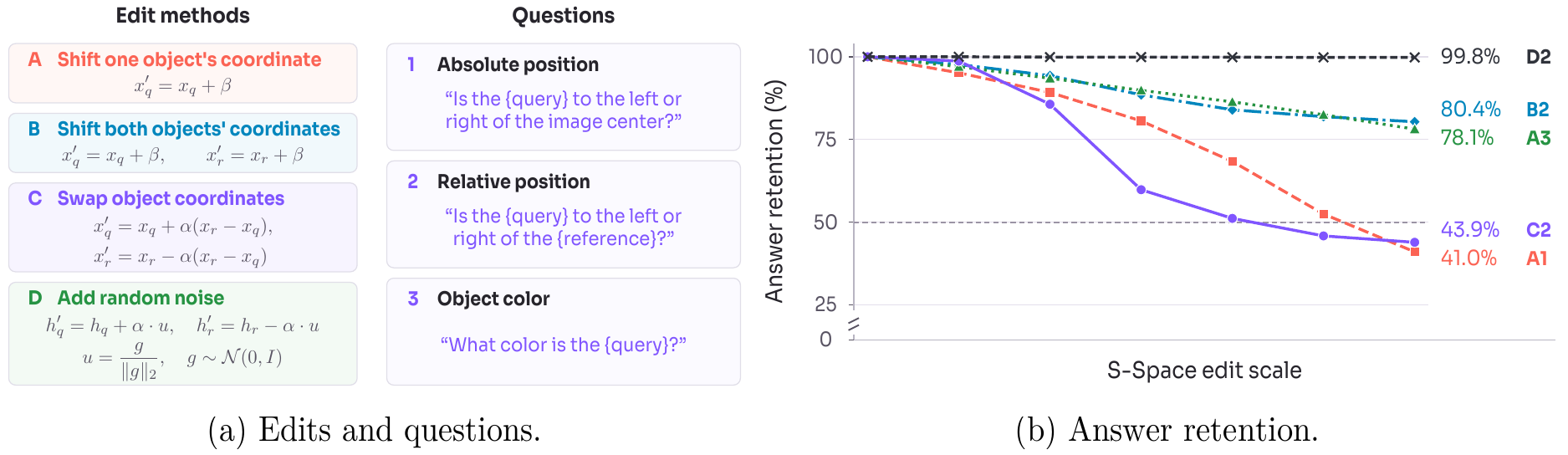}
  \vspace{-15pt}
  \caption{\textbf{Causal intervention of S-Space.} \textit{(a)} Different coordinate-editing and question configurations. \textit{(b)} Answer retention under different interventions and edit scales. Experiments use Molmo2-ER at layer 17 (L17). See Appendix~\ref{app:main-coordinate-interventions} for details.}
  \label{fig:sspace-7}
  \vspace{-6pt}
\end{figure}

\subsection{Accuracy}\label{section-accuracy}

We first examine how accurately S-Space represents object locations and their spatial relations.

\textbf{S-Space encodes object coordinates.} We project each object-token representation onto S-Space and compare the resulting coordinates with ground-truth object positions. As illustrated in Figure~\ref{fig:sspace-4}a and~\ref{fig:sspace-4}b, both individual object coordinates and pairwise coordinate differences strongly correlate with their ground-truth counterparts along the horizontal, vertical, and distance axes.

\textbf{S-Space supports spatial QA.} We next answer spatial-relation questions using only the coordinates read from S-Space, without relying on the model's verbal response, following Section~\ref{sec:method-readout}. As shown in Figure~\ref{fig:spatial-qa}c, this simple coordinate-based readout achieves accuracy comparable to the models' direct verbal answers. Moreover, it generalizes across CV-Bench, EmbSpatial-Bench, and SpatialTunnel, including benchmarks outside the distribution used to identify it.

Together, these results show that S-Space \textbf{encodes} spatial information with high fidelity.

\subsection{Causality}\label{section-causality}

Accurate readout alone does not necessarily imply that the model itself uses S-Space. To test its causal role, we read out object coordinates from S-Space, edit these coordinates, write the modified values back into the hidden representations, and then resume generation while leaving the image, question, and representation components outside S-Space unchanged.

\textbf{Editing S-Space coordinates changes spatial judgments.} As shown in Figure~\ref{fig:sspace-7}, S-Space intervention by coordinate editing substantially alter spatial judgments compared with random editing. The only exception is relative questions under common shifts, where shifting both objects together preserves their relative positions and therefore leaves the judgment merely unchanged.

\textbf{Non-spatial judgments are largely preserved.} The same intervention that strongly alters spatial answers have little effect on unrelated color judgments. This effect suggests that editing S-Space selectively modifies spatial information without broadly disrupting the object representation. Together, these results show that S-Space is causally and selectively \textbf{used} when the model produces verbal spatial judgments.

\vspace{-12pt}
\subsection{Persistence}\label{section-persistence}
\enlargethispage{\baselineskip}

Beyond the dataset generalization shown in Section~\ref{section-accuracy}, we further replace the relative-comparison prompts with independent single-object queries and even non-spatial questions that ask about colors. Object coordinates are read out in the same way and compared for relative spatial QA.

\begin{figure}[t]
  \centering
  \includegraphics[width=\linewidth]{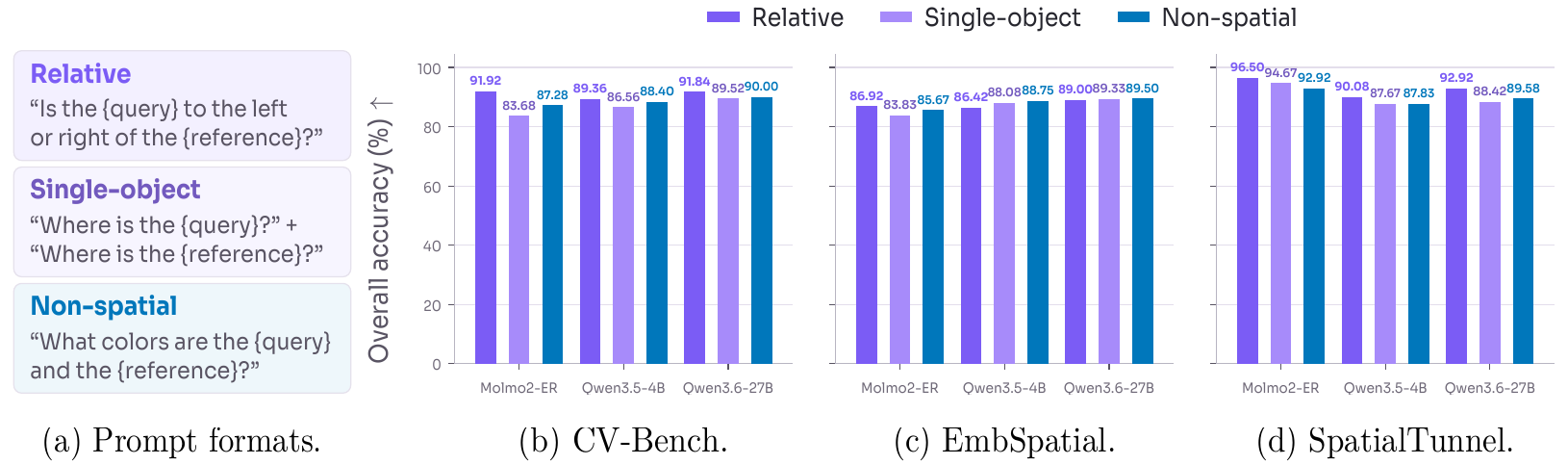}
  \vspace{-15pt}
  \caption{\textbf{S-Space persists across prompts.} The same S-Space readout continues to support accurate spatial judgments on CV-Bench, EmbSpatial, and SpatialTunnel \textit{(b-d)} across pairwise spatial, single-object, and non-spatial prompts \textit{(a)}.}
  \label{tab:prompt-robustness}
\end{figure}

\textbf{S-Space persists without explicit spatial questions.} As shown in Figure~\ref{tab:prompt-robustness}, despite substantial change in prompt structure and computational context, the resulting coordinates continue to support accurate spatial judgments. This robustness suggests that S-Space is an \textbf{intrinsic and persistent organization} of the model's representations, rather than an artifact induced by a particular dataset, prompt template, or pairwise-query format.

\section{How Does S-Space Emerge? A Case Study}\label{section-case-studies}

\begin{takeaway}
S-Space is hypothesized to emerge from predicting spatial language, which also endows it with broader structure beyond \textbf{visible objects} and \textbf{physical space}.
\end{takeaway}

We hypothesize that S-Space emerges from the objective of predicting spatial language during large-scale training. To predict tokens such as \textit{left} and \textit{right}, an MLLM must bind visible entities to their relative positions and make this information accessible to language generation. This training pressure may organize internal activations into stable spatial axes without explicit coordinate supervision. Intriguingly, the resulting structure inherited from language may extend beyond visible objects and physical space, as illustrated by the following showcases.

\begin{figure}[t]
  \centering
  \includegraphics[width=\linewidth]{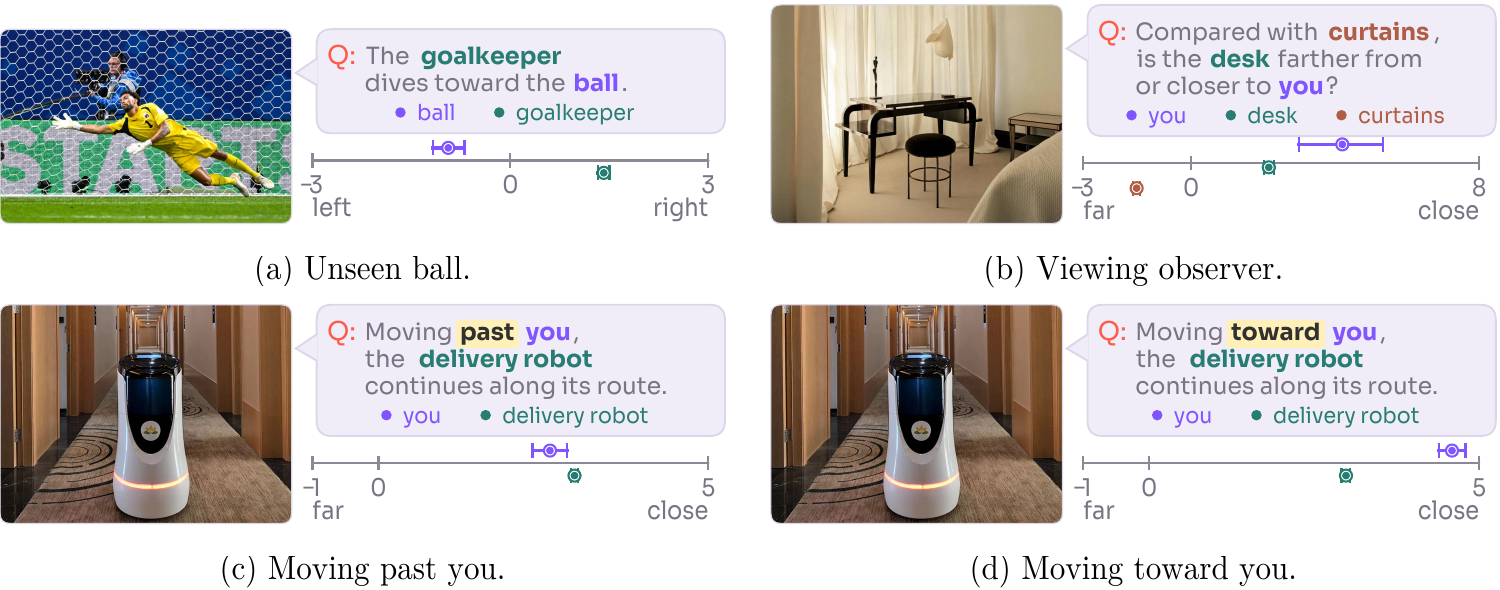}
  \vspace{-20pt}
  \caption{\textbf{S-Space beyond visible objects.} It localizes off-screen entities \textit{(a)}, places the invisible observer \textit{(b)}, and reflects relations implied by language \textit{(c-d)}. Readout from Molmo2-ER L17. Bars show 95\% prompt-template bootstrap intervals.}
  \label{fig:case-beyond}\label{fig:beyond-visible-selected}
\end{figure}

\begin{figure}[t]
  \centering
  \includegraphics[width=1.00\linewidth]{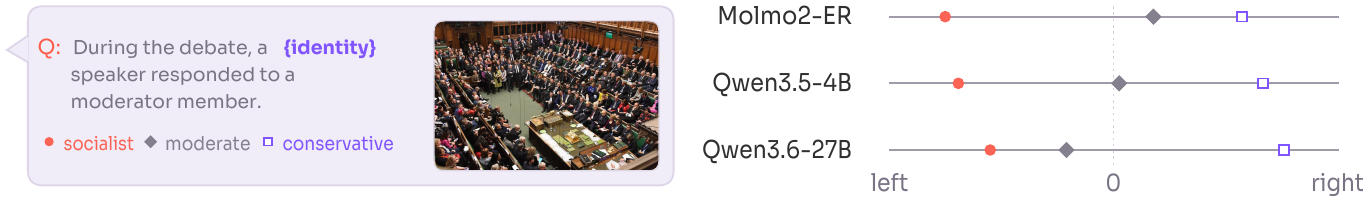}
  \vspace{-15pt}
  \caption{\textbf{S-Space beyond physical space}. \textit{Socialist} and \textit{conservative} project toward opposite ends of the horizontal axis of S-Space, mirroring the linguistic metaphor of political left and right.}
  \label{fig:case-language}\label{fig:sspace-11}
\end{figure}

\paragraph{S-Space reflects relations implied by visual evidence.}\label{section-beyond-the-visible}
Even when an entity itself is not visible, its position in S-Space can be inferred from visual evidence. As shown in Figure~\ref{fig:case-beyond}a, a goalkeeper diving toward an unseen ball places the token \textit{ball} in the direction of the dive. Beyond unseen objects, even the observer, represented by the abstract token \emph{you}, projects toward the \emph{close} end of the distance axis (Figure~\ref{fig:case-beyond}b). These cases suggest that S-Space does not merely register detected objects, but can complete the surrounding spatial world.

\paragraph{S-Space reflects relations implied by language.}\label{section-beyond-text}
Linguistic context can likewise determine where an entity is represented in S-Space. As shown in Figure~\ref{fig:case-beyond}c and~\ref{fig:case-beyond}d, we keep the delivery-robot image fixed while replacing \emph{moving past you} with \emph{moving toward you}. Correspondingly, the token \emph{you} shifts from a farther to a closer distance coordinate relative to the robot, while the robot's own coordinate remains nearly unchanged. 

\paragraph{S-Space blends physical and abstract concepts.}
\label{section-language-supervision-blends-physical-and-abstract-concepts}
Language is a highly abstract representation of the world: spatial terms describe physical geometry, but are also routinely repurposed to express abstract meanings. Models trained on language may therefore inherit associations that do not cleanly separate physical positions from their abstract uses. Strikingly, as illustrated in Figure~\ref{fig:case-language}, \emph{socialist}---a concept conventionally associated with the political left---also projects toward the left end of the horizontal axis, and \emph{conservative} toward the right. These cases illustrate both the capabilities and limitations of spatial structure inherited from large-scale language training.

\section{Diagnosing Spatial Reasoning Failures with S-Space}\label{section-is-s-space-proficiently-used-by-internal-reasoning}

\begin{takeaway}
MLLMs often possess sufficiently accurate egocentric \textbf{spatial representations} in S-Space, yet still fail in downstream spatial reasoning. These failures reveal two key bottlenecks: (1) imperfect \textbf{spatial computation} over accurate representations (\S\ref{section-failure-analysis-in-spatial-reasoning}), and (2) \textbf{reference-frame coupling} when reconciling different coordinate systems (\S\ref{section-nsew-case}).
\end{takeaway}

Modern MLLMs still struggle with spatial reasoning, especially when they must build a coherent spatial understanding from partial viewpoints. This requires correctly transforming between viewpoints and reconciling different coordinate frames. S-Space offers a direct view of the model's internal spatial structure, enabling us to diagnose where and how these reasoning failures arise.

\subsection{Perception-Computation Decomposition via S-Space}\label{section-failure-analysis-in-spatial-reasoning}

Perspective-taking questions require judging object relations after an imagined change of viewpoint. We use SpinBench~\citep{zhang2026spinbench} as our evaluation benchmark, which asks which of two objects would appear to the left, or closer to the observer, from a left, right, or back view. Yet frontier VLMs are reported to perform surprisingly poorly on these seemingly simple transformations. This raises a natural question: does the failure arise from spatial perception, or from the computation performed to transform the perceived spatial representation?

\begin{figure}[t]
  \centering
  \includegraphics[width=\linewidth]{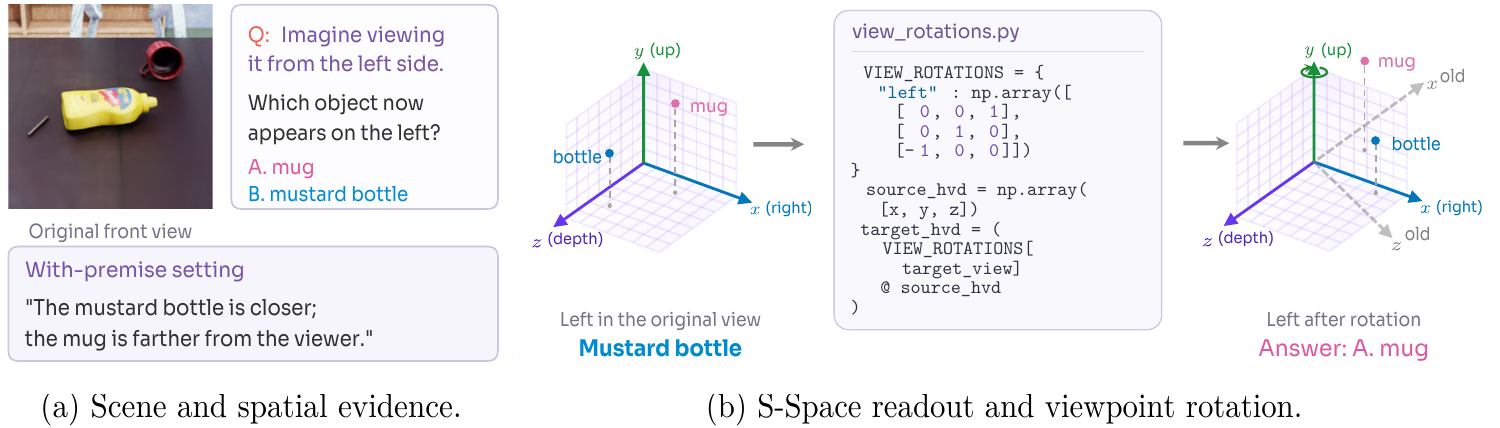}
  \vspace{-15pt}
  \caption{\textbf{Perception-Computation decomposition via S-Space.} \textit{(a)} Perspective-taking question and the optional w/ premise setting. \textit{(b)} We separate spatial perception from spatial computation by reading object coordinates from S-Space, applying the viewpoint transformation explicitly, and deriving the answer with a deterministic spatial rule.}
  \label{fig:sspace-14}
\end{figure}

S-Space allows us to disentangle these two sources of error by reading the model's internal coordinates independently of its final answer. As shown in Figure~\ref{fig:sspace-14}, we construct a decomposed pipeline that reads the objects' coordinates from S-Space, applies a fixed rotation corresponding to the target viewpoint, and derives the answer with the same deterministic spatial rule as in Section~\ref{sec:method-readout}. We compare this method with the model's native answer generation, both with and without chain-of-thought reasoning \citep{wei2022chain}.

\begin{table}[t]
  \centering
  \small
  \setlength{\tabcolsep}{2.2pt}
     \begin{tabular*}{\linewidth}{@{\extracolsep{\fill}}lcccccc@{}}
      \toprule
      & \multicolumn{3}{c}{\mirrosth{w/o premise}}
      & \multicolumn{3}{c}{\mirrosth{w/ premise}} \\
      \cmidrule(lr){2-4}
      \cmidrule(lr){5-7}
    
      \mirrosth{Model}
      & \mirrosth{Direct}
      & \mirrosth{Reasoning}
      & \mirrosth{\shortstack{S-Space + Rotation}}
      & \mirrosth{Direct}
      & \mirrosth{Reasoning}
      & \mirrosth{\shortstack{S-Space + Rotation}} \\
      \midrule
    
Molmo2-ER & 60.96 & --- & \textbf{94.52} (+33.56) & 61.81 & --- & \textbf{99.31} (+37.50) \\
Qwen3.5-4B & 60.96 & 62.33 (+1.37) & \textbf{91.10} (+30.14) & 60.42 & 72.22 (+11.80) & \textbf{97.92} (+37.50) \\
Qwen3.6-27B & 60.96 & 83.56 (+22.60) & \textbf{95.21} (+34.25) & 72.92 & 91.67 (+18.75) & \textbf{99.31} (+26.39)%
\\
    
      \bottomrule
    \end{tabular*}
  \caption{\textbf{SpinBench perspective-taking accuracy.} Numbers in parentheses show gains over direct answers in the same setting, in percentage points; dashes indicate no reasoning mode.}
  \label{tab:spinbench}
  \label{tab:spinbench-premise}
\end{table}

\paragraph{Explicit transformation over S-Space substantially improves perspective-taking.}
As shown in Table~\ref{tab:spinbench}, direct generation remains limited across all three models. Chain-of-thought improves performance with additional test-time computation, but notable errors persist. In contrast, the prespecified S-Space readout followed by explicit coordinate transformation yields consistently stronger results, which shows that the models encode sufficient spatial information for accurate judgments.

\paragraph{Reasoning errors persist with explicit spatial premises.} Following the premise setting in SpinBench, we further provide the correct source-view relation serves as an oracle for spatial perception, bypassing the model’s own perceptual inference (Figure~\ref{fig:sspace-14}a). Even with this relation supplied, models often fail to produce the correct target-view answer. Meanwhile, their S-Space representations become more accurate with language implication, consistent with Section~\ref{section-case-studies}, and support near-perfect accuracy. This further highlights a key bottleneck in reliably transforming and operating on correct spatial information during viewpoint reasoning.

\subsection{Tracking S-Space Transformation through Reasoning}\label{section-computation-scaling-transforming-s-space-through-reasoning}

\begin{figure}[b]
  \centering
  \vspace{-10pt}
  \includegraphics[width=\linewidth]{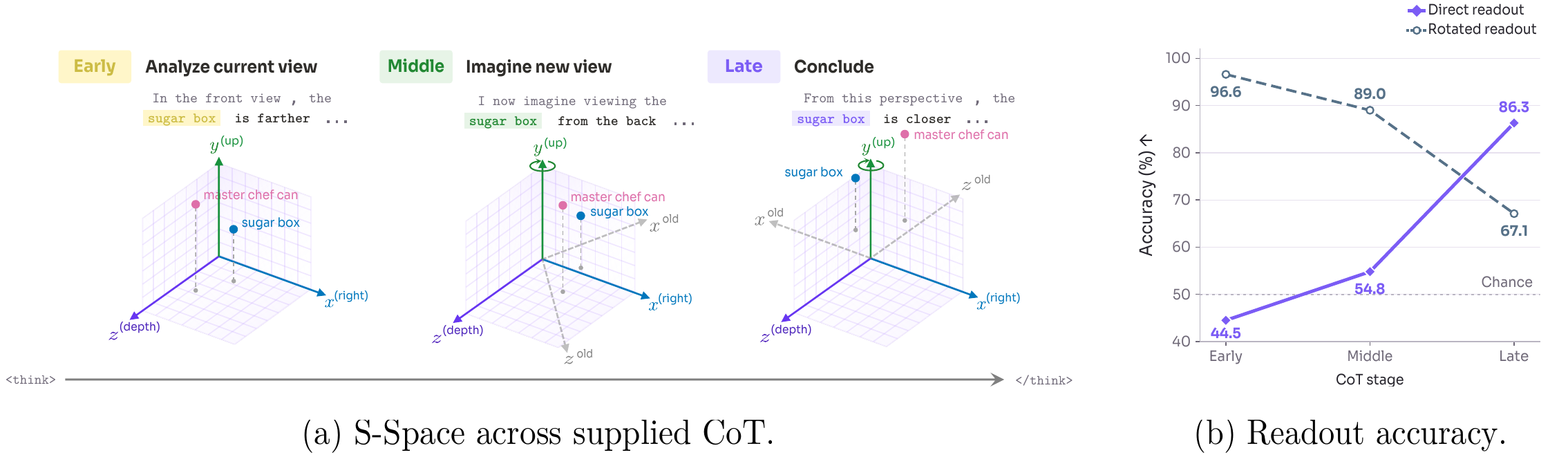}
  \vspace{-15pt}
  \caption{\textbf{Chain-of-thought progressively transforms S-Space.} \textit{(a)} S-Space coordinates read out from object names at successive stages of a controlled CoT. \textit{(b)} Qwen3.6-27B L43 readout accuracy on SpinBench with and without explicit viewpoint rotation.
  }
  \label{fig:sspace-15}
\end{figure}

Although still imperfect, chain-of-thought reasoning indeed improves over direct answer generation. We therefore track object coordinates in S-Space throughout the chain of thought to examine how these spatial representations change during the model's own reasoning process. This analysis complements the internal-workspace perspective in Section~\ref{section-the-spatial-world-in-s-space} by revealing internal \textbf{transformation} of spatial information in S-Space.

We conduct a controlled analysis on Qwen3.6-27B. Specifically, as shown in Figure~\ref{fig:sspace-15}a, we manually construct a CoT template that explicitly identifies the original spatial relation, imagines the target viewpoint, and derives the transformed relation. By probing repeated object mentions across these stages, we evaluate how S-Space updates accordingly. At each stage, we read out the S-Space coordinates and use them to answer the corresponding SpinBench perspective-taking question, either directly or after applying the optimal viewpoint rotation.

\vspace{-5pt}
\paragraph{Reasoning progressively transforms S-Space.} 
As the chain-of-thought unfolds, the objects' relative coordinates become increasingly aligned with the viewpoint requested by the question. This transformation is reflected in a progressive shift from an advantage for the rotated readout at earlier stages to an advantage for the direct readout at later stages, as shown in Figure~\ref{fig:sspace-15}b. This suggests that the model internally simulates part of the required perspective transformation. However, this transformation remains imperfect and does not always produce the correct verbal answer.

\FloatBarrier
\subsection{Diagnosing Reference-Frame Coupling in S-Space}\label{section-nsew-case}

S-Space also helps interpret other reasoning errors. In cardinal-direction spatial reasoning, we repeatedly observe reference-frame coupling: models often infer cardinal directions from image-relative relations, mapping \emph{right} to \emph{east} or \emph{above} to \emph{north}, as showcased in Figure~\ref{fig:cardinal-diagnosis}a. To investigate this phenomenon, using positional relationship tasks of MMSI-Bench~\citep{yang2026mmsi}, we collect cardinal-direction tokens (\textit{north}, \textit{south}, etc.) from Qwen3.6-27B's chain-of-thought and read out their coordinates in S-Space. 
\vspace{-5pt}
\paragraph{Cardinal and image-relative directions are coupled in S-Space.}
As shown in Figure~\ref{fig:cardinal-diagnosis}b, the resulting geometry makes the coupling explicit: \emph{east} aligns with right in S-Space, \emph{north} with above, and \textit{northeast} reflects both directions. This suggests that the model represents cardinal and image-relative directions through partially shared spatial features rather than fully independent concepts. Inspired by Section~\ref{section-case-studies}, we hypothesize that this coupling may partly reflect biases in web-scale pretraining data, where cardinal directions frequently appear in map-like reference frames aligned with horizontal and vertical axes.

\begin{center}
  \includegraphics[width=0.90\linewidth]{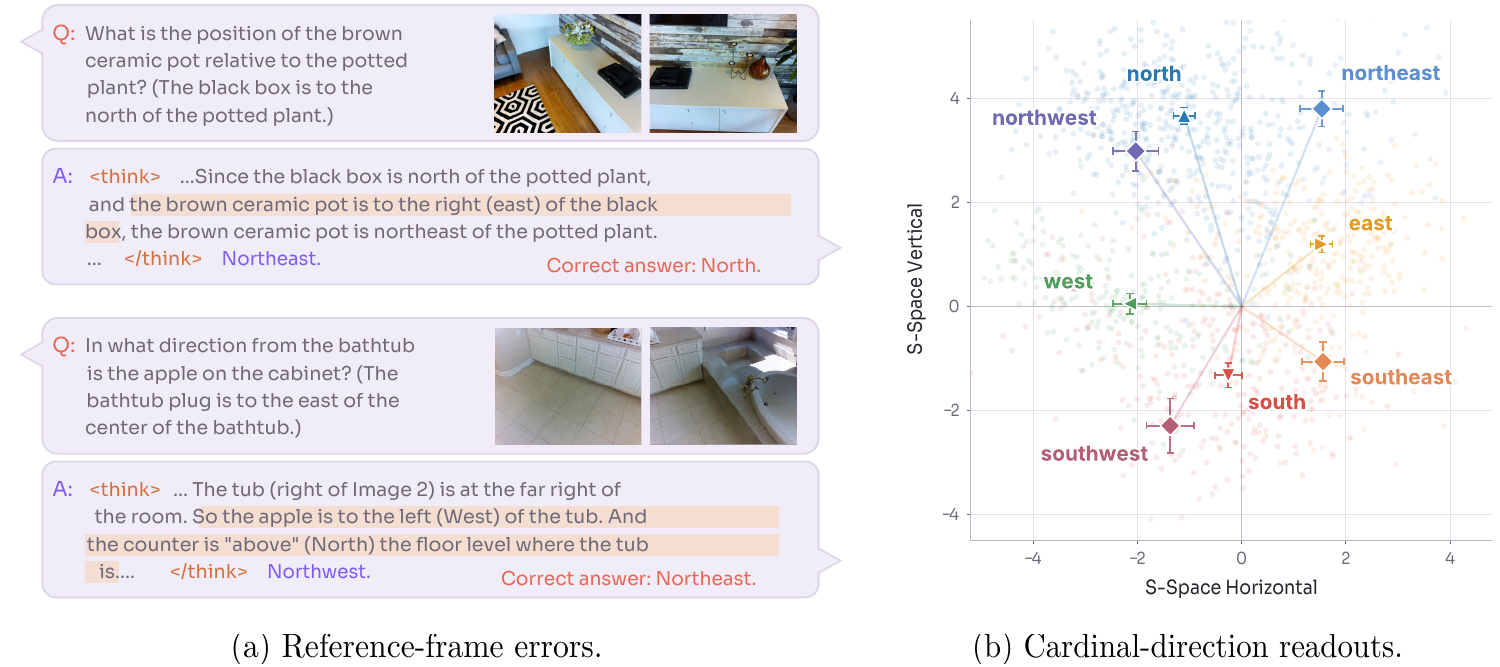}
  \vspace{-10pt}
  \captionof{figure}{\textbf{Direction coupling in S-Space.} \textit{(a)} Representative reasoning failures where image-relative relations are incorrectly reused as cardinal directions. \textit{(b)} S-Space coordinates of cardinal-direction tokens from Qwen3.6-27B chain-of-thought on MMSI-Bench.}
  \label{fig:cardinal-diagnosis}
  \label{fig:sspace-12}
\end{center}

\section{Discussions}\label{section-conclusion}\label{section-what-s-next}

We identified S-Space as an internal spatial workspace whose contents can be measured and operated on to investigate spatial reasoning failures. More broadly, such mechanistic analysis (see related work in Appendix~\ref{section-related-work}) helps reveal which key capabilities do and do not emerge through scaling, and for capabilities that remain unreliable, diagnose the missing ingredients. Our results have uncovered biases inherited from pre-training. Another implication is that future systems need not rely on a monolithic MLLM to perform every spatial operation internally. Instead, agentic spatial reasoning~\citep{cho2026spatialclaw, luo2026pyspatial} can connect dedicated tools for transformation, simulation, and memory, ultimately turning capable perception into reliable reasoning and action.

\section*{Authors}
Zhiyuan Shi\textsuperscript{*}, Xiang Li\textsuperscript{*}, Xingzhuo Guo\textsuperscript{*}, Shangchen Miao, Weiliang Chen, Qiyu Dai, Jinshan Ren, Jiangran Lyu, Fangfu Liu, Mingsheng Long, Jialong Wu\textsuperscript{\textdagger}

\section*{Affiliations}
MirroS, Tsinghua University, Peking University

\begingroup
\renewcommand{\thefootnote}{\fnsymbol{footnote}}
\footnotetext[1]{Equal contribution.}
\footnotetext[2]{Project lead.}
\endgroup

\printbibliography[heading=bibintoc,title={References}]\label{section-references}

\clearpage
\appendix
\input{appendix_style.sty}

\section{Related Work}\label{section-related-work}

\paragraph{Evaluating spatial reasoning.} Benchmarks for evaluating the spatial capabilities of MLLMs have become increasingly challenging and comprehensive. Early efforts focused primarily on single-image spatial understanding~\citep{tong2024cambrian, du2024embspatial}, including relative position reasoning and object size estimation. Subsequent studies extended evaluation to multi-view~\citep{yang2026mmsi, yin2025spatial} and video~\citep{yang2025thinking,lin2025mmsi,zhang2026spinbench} settings, requiring models to integrate observations into coherent spatial representations. These spatial capabilities have been incorporated into embodied tasks involving grounding~\citep{yuan2024robopoint, team2025gemini} and active exploration~\citep{hong2026esi, zhang2026theory}. Nevertheless, spatial reasoning remains a fundamental challenge for MLLMs: visual spatial puzzles~\citep{chen2026babyvision, wang2025spatialviz} continue to reveal substantial performance gaps relative to humans, motivating deeper diagnosis and mechanistic interpretation of these limitations.

\paragraph{Mechanistic interpretability.} Internal representation in LLMs are hypothesized to composed linearly~\citep{park2024linear, elhage2022superposition}. Sparse autoencoders (SAEs) learn dictionaries that recover interpretable features from model activations~\cite{huben2024sparse, bricken2023monosemanticity, templeton2024scaling}. Moving from representation to computation, transcoders approximate MLP computations with sparse features, enabling circuit tracing to identify interactions underlying intermediate reasoning steps~\citep{dunefsky2024transcoders, ameisen2025circuit}. Most closely related to our work, J-Space~\citep{gurnee2026verbalizable} introduces the Jacobian lens,  which uses Jacobians to map intermediate activations into the output vocabulary, thereby verbalizing internal representations. Applying this lens across layers reveals internal concepts and intermediate computations. S-Space extends this Jacobian-based perspective to spatial representations in MLLMs, focusing on continuous spatial coordinates rather than discrete verbalizable concepts to investigate how models internally represent and reason about space.

\paragraph{Interpreting spatial representations.} MLLMs bind spatial information from visual inputs to the textual representations of object names, enabling both spatial readout and causal intervention~\citep{kang2026linear}. Prior work have examined spatial binding~\citep{cui2026dual}, the functional role of attention~\citep{chen2025spatial, fu2026multimodal}, and bias such as vertical-depth coupling~\citep{min2026whyfar}. Much of this work focuses on synthetic settings and two-dimensional pairwise relations. S-Space operates on real-world images and three-dimensional spatial coordinates, providing a simple, unified interface for spatial information readout, causal intervention, and tracking viewpoint transformation.

\section{Theoretical Insights of S-Space}
\label{section-theory}

We now formalize the intuition underlying our procedure for identifying S-Space.

Let \(\mathcal S^\star\) denote the latent three-dimensional spatial subspace hypothesized in Section~\ref{sec:method-formulation}, where \(u_x,u_y,u_z\) are unit directions corresponding to the horizontal, vertical, and distance axes. Our procedure in Section~\ref{sec:method-identification} produces estimated directions \(v_x,v_y,v_z\). Below, we give sufficient conditions under which these estimated directions recover the corresponding latent spatial directions.

Following the linear representation hypothesis~\citep{park2024linear,kang2026linear}, we assume that distinct types of semantic information are linearly composed within hidden representations. Since object-token representations encode both spatial and non-spatial information, recovering spatial directions requires separating their respective contributions. We introduce the following assumptions to characterize this structure.

\begin{assumption}[Representation decomposition]
At an object-token position \(t\), the hidden state can be conceptually decomposed into a spatial component and two non-spatial:
\begin{equation}\label{eq:method-state-decomposition}
\begin{gathered}
h_t=h_t^{\mathcal S^\star}+h_t^{\mathrm{non\text{-}spatial}}+h_t^{\mathrm{context}},\\
h_t^{\mathcal S^\star}\in\mathcal S^\star,\qquad
h_t^{\mathrm{non\text{-}spatial}},h_t^{\mathrm{context}}\in(\mathcal S^\star)^\perp.
\end{gathered}
\end{equation}
\end{assumption}
Here, \(h_t^{\mathcal S^\star}\) encodes the object's viewer-centered spatial attributes; \(h_t^{\mathrm{non\text{-}spatial}}\) captures its non-spatial attributes, such as identity, appearance, and function; and \(h_t^{\mathrm{context}}\) captures information contributed by the question semantics, prompt phrasing, and broader scene context.

\begin{assumption}[Gradient decomposition]
Denote $d_{p,t,g}=\nabla_{h_t}s_{p,g}$, the gradient are analogously decomposed into components associated with spatial, non-spatial, and contextual information:
\begin{equation}\label{eq:method-jacobian-decomposition}
d_{p,t,g}
=
d_{p,t,g}^{\mathcal S^\star}
+d_{p,t,g}^{\mathrm{non\text{-}spatial}}
+d_{p,t,g}^{\mathrm{context}}.
\end{equation}
\end{assumption}

\begin{assumption}[Gradient behavior]
For a relative spatial question along axis \(g\), the spatial component of the answer-contrast is aligned with the corresponding latent direction \(u_g\) and changes sign between the query and reference:
\begin{equation}\label{eq:method-component-behavior}
d_{p,q,g}^{\mathcal S^\star}
=
\beta_{p,g}u_g,\qquad
d_{p,r,g}^{\mathcal S^\star}
=
-\beta_{p,g}u_g,
\end{equation}
where \(\mathbb E_p[\beta_{p,g}]>0\). Meanwhile:
\begin{equation}
d_{p,q,g}^{\mathrm{non\text{-}spatial}}=d_{p,r,g}^{\mathrm{non\text{-}spatial}}=0,\quad
\mathbb E_p\!\left[d_{p,q,g}^{\mathrm{context}}-d_{p,r,g}^{\mathrm{context}}\right]=0.
\end{equation}
\end{assumption}
These conditions capture three intuitions. First, the spatial contributions have opposite signs for the query and reference. Second, non-spatial object attributes do not affect the answer contrast. Third, role-asymmetric contextual effects may vary across examples but cancel in expectation. Together, they motivate antisymmetrizing the two object gradients and averaging across prompts and examples.

\begin{property}
Under Assumptions 1--3, role antisymmetrization recovers the latent spatial direction \(u_g\) in expectation:
\begin{equation}\label{eq:method-spatial-expectation}
\mathbb E_p\!\left[a_{p,g}\right]
=
\mathbb E_p[\beta_{p,g}]\,u_g.
\end{equation}
Consequently, the estimator in Eq.~(\ref{eq:method-antisymmetric-jacobian}) satisfies $v_g=u_g$, up to the choice of orientation, and the recovered subspace
\(\mathcal S=\operatorname{span}\{v_x,v_y,v_z\}\)
matches the latent spatial subspace \(\mathcal S^\star\).
\end{property}

\begin{proof}
Substituting the gradient decomposition into Eq.~(\ref{eq:method-antisymmetric-jacobian}) and applying Eq.~(\ref{eq:method-component-behavior}), we obtain
\begin{equation}\label{eq:method-spatial-proof}
\begin{aligned}
a_{p,g}
&=\frac{1}{2}\left(
d_{p,q,g}^{\mathcal S^\star}
-d_{p,r,g}^{\mathcal S^\star}
\right)+\frac{1}{2}\left(
d_{p,q,g}^{\mathrm{non\text{-}spatial}}
-d_{p,r,g}^{\mathrm{non\text{-}spatial}}
\right)\\
&\quad+\frac{1}{2}\left(
d_{p,q,g}^{\mathrm{context}}
-d_{p,r,g}^{\mathrm{context}}
\right)\\
&=\beta_{p,g}u_g
+\frac{1}{2}\left(
d_{p,q,g}^{\mathrm{context}}
-d_{p,r,g}^{\mathrm{context}}
\right).
\end{aligned}
\end{equation}
The non-spatial terms vanish. Taking expectations eliminates the remaining contextual difference, yielding Eq.~(\ref{eq:method-spatial-expectation}). Since \(\mathbb E_p[\beta_{p,g}]>0\), normalization recovers \(u_g\).
\end{proof}

\section{Implementation Details and Extended Results}\label{section-algorithm}\label{section-implementation-details}

\subsection{Identifying S-Space}\label{app:data}

\paragraph{Models and data.}
We evaluate Molmo2-ER, MolmoAct2-Pretrain, MolmoAct2~\citep{fang2026molmoact2}, Qwen3.5-4B, and Qwen3.6-27B~\citep{qwen2026qwen35,qwen2026qwen36}. MolmoAct2-Pretrain and MolmoAct2 are VLA models derived from Molmo2-ER after successive stages of action training, allowing us to examine how spatial representations change across these stages. Axis identification and layer selection use disjoint sets of 6,000 and 1,800 COCO-2017 training images, with one object pair per image. For each of the six spatial relations (left, right, above, below, far, and close), we use 1,000 examples to identify the axes and 300 examples to select the layer. The supplementary code provides the image IDs and object pair records for both sets.

We obtain each object's ground truth horizontal and vertical positions from the center of its annotated bounding box, divided by the image width and height, respectively. For depth, we average the inverse depth predicted by Depth Anything V2-Small~\citep{yang2024depthanythingv2} over pixels within each COCO instance mask. Larger values indicate closer objects.

\paragraph{Identification prompts.}\label{section-identification-prompts}

We use the five templates below for each spatial axis. For each object pair, we fill every template twice, exchanging which object is named in \{query\} and which is named in \{reference\}. We compute gradients with respect to the hidden state at the token that ends each object name in the prompt.

\appendixprompt{Horizontal: left / right}{%
\numberedprompt{1}{Is the \{query\} to the left or right of the \{reference\}? Answer with left or right.}
\numberedprompt{2}{Where is the \{query\} relative to the \{reference\}: left or right? Answer with left or right.}
\numberedprompt{3}{In the image, is the \{query\} to the left or right of the \{reference\}? Answer with left or right.}
\numberedprompt{4}{Which describes the position of the \{query\} relative to the \{reference\}: left or right? Answer with left or right.}
\numberedprompt{5}{Relative to the \{reference\}, is the \{query\} on the left or right? Answer with left or right.}
}
\appendixprompt{Vertical: above / below}{%
\numberedprompt{1}{Is the \{query\} above or below the \{reference\}? Answer with above or below.}
\numberedprompt{2}{Where is the \{query\} relative to the \{reference\}: above or below? Answer with above or below.}
\numberedprompt{3}{In the image, is the \{query\} above or below the \{reference\}? Answer with above or below.}
\numberedprompt{4}{Which describes the position of the \{query\} relative to the \{reference\}: above or below? Answer with above or below.}
\numberedprompt{5}{Relative to the \{reference\}, is the \{query\} above or below? Answer with above or below.}
}
\appendixprompt{Distance: far / close}{%
\numberedprompt{1}{Is the \{query\} farther from or closer to the camera than the \{reference\}? Answer with far or close.}
\numberedprompt{2}{Considering only depth, is the \{query\} farther from or closer to the camera than the \{reference\}? Answer far or close.}
\numberedprompt{3}{In the image, is the \{query\} farther from or closer to the camera than the \{reference\}? Answer with far or close.}
\numberedprompt{4}{Choose far or close: is the \{query\} farther from or closer to the camera than the \{reference\}? Answer with far or close.}
\numberedprompt{5}{Is the \{query\} farther from the camera than the \{reference\}, or closer? Answer far or close.}
}

\paragraph{Layer selection.}
We evaluate validation accuracy for each layer on the 1,800 relative spatial question examples constructed from COCO. For each example, we average S-Space coordinate differences across the five prompt templates for spatial judgments. We select the layer with the highest overall accuracy across the three axes. Table~\ref{tab:spatial-qa} lists the selected layers, indexed from zero after each transformer block. We use these layers unless specified otherwise.

We report S-Space accuracy at every evaluated layer on the COCO selection set in Table~\ref{tab:layer-sweep}.

\begin{table}[!t]
\begingroup
\colorlet{SweepPurple}{MirrosAccent}
\setlength{\tabcolsep}{4.3pt}
\renewcommand{\arraystretch}{0.90}
\fontsize{9}{9.3}\selectfont
\centering

\begin{tabular}{@{}rrrrrrrrrrrr@{}}
\toprule
\multicolumn{2}{c}{\mirrosth{Molmo2-ER}}
& \multicolumn{2}{c}{\mirrosth{\shortstack{MolmoAct2\\Pretrain}}}
& \multicolumn{2}{c}{\mirrosth{MolmoAct2}}
& \multicolumn{2}{c}{\mirrosth{Qwen3.5-4B}}
& \multicolumn{4}{c}{\mirrosth{Qwen3.6-27B}} \\
\cmidrule(lr){1-2}
\cmidrule(lr){3-4}
\cmidrule(lr){5-6}
\cmidrule(lr){7-8}
\cmidrule(lr){9-12}

\mirrosth{Layer} & \mirrosth{Acc.}
& \mirrosth{Layer} & \mirrosth{Acc.}
& \mirrosth{Layer} & \mirrosth{Acc.}
& \mirrosth{Layer} & \mirrosth{Acc.}
& \multicolumn{2}{c}{\mirrosth{Layers 0--31}}
& \multicolumn{2}{c}{\mirrosth{Layers 32--62}} \\
\cmidrule(lr){9-10}
\cmidrule(lr){11-12}

& & & & & & & &
\mirrosth{Layer} & \mirrosth{Acc.} & \mirrosth{Layer} & \mirrosth{Acc.} \\
\midrule

0 & \cellcolor{SweepPurple!4.65}55.17
& 0 & \cellcolor{SweepPurple!6.90}57.67
& 0 & \cellcolor{SweepPurple!5.65}56.28
& 0 & \cellcolor{SweepPurple!2.35}52.61
& 0 & \cellcolor{SweepPurple!7.40}58.22
& 32 & \cellcolor{SweepPurple!34.65}88.50 \\

1 & \cellcolor{SweepPurple!4.90}55.44
& 1 & \cellcolor{SweepPurple!7.85}58.72
& 1 & \cellcolor{SweepPurple!7.70}58.56
& 1 & \cellcolor{SweepPurple!2.25}52.50
& 1 & \cellcolor{SweepPurple!6.40}57.11
& 33 & \cellcolor{SweepPurple!33.55}87.28 \\

2 & \cellcolor{SweepPurple!6.05}56.72
& 2 & \cellcolor{SweepPurple!8.40}59.33
& 2 & \cellcolor{SweepPurple!7.15}57.94
& 2 & \cellcolor{SweepPurple!4.05}54.50
& 2 & \cellcolor{SweepPurple!6.90}57.67
& 34 & \cellcolor{SweepPurple!33.30}87.00 \\

3 & \cellcolor{SweepPurple!6.40}57.11
& 3 & \cellcolor{SweepPurple!8.85}59.83
& 3 & \cellcolor{SweepPurple!7.25}58.06
& 3 & \cellcolor{SweepPurple!5.70}56.33
& 3 & \cellcolor{SweepPurple!5.85}56.50
& 35 & \cellcolor{SweepPurple!36.70}90.78 \\

4 & \cellcolor{SweepPurple!6.60}57.33
& 4 & \cellcolor{SweepPurple!8.55}59.50
& 4 & \cellcolor{SweepPurple!8.55}59.50
& 4 & \cellcolor{SweepPurple!6.20}56.89
& 4 & \cellcolor{SweepPurple!6.20}56.89
& 36 & \cellcolor{SweepPurple!36.20}90.22 \\

5 & \cellcolor{SweepPurple!5.30}55.89
& 5 & \cellcolor{SweepPurple!9.85}60.94
& 5 & \cellcolor{SweepPurple!9.15}60.17
& 5 & \cellcolor{SweepPurple!5.70}56.33
& 5 & \cellcolor{SweepPurple!6.70}57.44
& 37 & \cellcolor{SweepPurple!36.20}90.22 \\

6 & \cellcolor{SweepPurple!6.80}57.56
& 6 & \cellcolor{SweepPurple!10.25}61.39
& 6 & \cellcolor{SweepPurple!9.90}61.00
& 6 & \cellcolor{SweepPurple!9.15}60.17
& 6 & \cellcolor{SweepPurple!6.90}57.67
& 38 & \cellcolor{SweepPurple!35.55}89.50 \\

7 & \cellcolor{SweepPurple!7.25}58.06
& 7 & \cellcolor{SweepPurple!10.95}62.17
& 7 & \cellcolor{SweepPurple!9.20}60.22
& 7 & \cellcolor{SweepPurple!24.55}77.28
& 7 & \cellcolor{SweepPurple!7.15}57.94
& 39 & \cellcolor{SweepPurple!36.05}90.06 \\

8 & \cellcolor{SweepPurple!7.30}58.11
& 8 & \cellcolor{SweepPurple!11.10}62.33
& 8 & \cellcolor{SweepPurple!9.35}60.39
& 8 & \cellcolor{SweepPurple!18.65}70.72
& 8 & \cellcolor{SweepPurple!8.35}59.28
& 40 & \cellcolor{SweepPurple!35.80}89.78 \\

9 & \cellcolor{SweepPurple!14.00}65.56
& 9 & \cellcolor{SweepPurple!14.70}66.33
& 9 & \cellcolor{SweepPurple!15.10}66.78
& 9 & \cellcolor{SweepPurple!14.10}65.67
& 9 & \cellcolor{SweepPurple!8.50}59.44
& 41 & \cellcolor{SweepPurple!34.05}87.83 \\

10 & \cellcolor{SweepPurple!12.70}64.11
& 10 & \cellcolor{SweepPurple!16.10}67.89
& 10 & \cellcolor{SweepPurple!15.30}67.00
& 10 & \cellcolor{SweepPurple!22.15}74.61
& 10 & \cellcolor{SweepPurple!8.90}59.89
& 42 & \cellcolor{SweepPurple!36.25}90.28 \\

11 & \cellcolor{SweepPurple!15.60}67.33
& 11 & \cellcolor{SweepPurple!17.85}69.83
& 11 & \cellcolor{SweepPurple!17.05}68.94
& 11 & \cellcolor{SweepPurple!34.15}87.94
& 11 & \cellcolor{SweepPurple!9.25}60.28
& 43 & \cellcolor{SweepPurple!37.60}\textbf{91.78} \\

12 & \cellcolor{SweepPurple!15.05}66.72
& 12 & \cellcolor{SweepPurple!16.55}68.39
& 12 & \cellcolor{SweepPurple!16.95}68.83
& 12 & \cellcolor{SweepPurple!28.75}81.94
& 12 & \cellcolor{SweepPurple!9.10}60.11
& 44 & \cellcolor{SweepPurple!36.65}90.72 \\

13 & \cellcolor{SweepPurple!27.40}80.44
& 13 & \cellcolor{SweepPurple!29.05}82.28
& 13 & \cellcolor{SweepPurple!28.65}81.83
& 13 & \cellcolor{SweepPurple!25.55}78.39
& 13 & \cellcolor{SweepPurple!8.35}59.28
& 45 & \cellcolor{SweepPurple!36.20}90.22 \\

14 & \cellcolor{SweepPurple!25.00}77.78
& 14 & \cellcolor{SweepPurple!27.40}80.44
& 14 & \cellcolor{SweepPurple!25.05}77.83
& 14 & \cellcolor{SweepPurple!27.35}80.39
& 14 & \cellcolor{SweepPurple!7.75}58.61
& 46 & \cellcolor{SweepPurple!36.25}90.28 \\

15 & \cellcolor{SweepPurple!29.55}82.83
& 15 & \cellcolor{SweepPurple!32.60}86.22
& 15 & \cellcolor{SweepPurple!30.60}84.00
& 15 & \cellcolor{SweepPurple!34.95}\textbf{88.83}
& 15 & \cellcolor{SweepPurple!13.20}64.67
& 47 & \cellcolor{SweepPurple!34.45}88.28 \\

16 & \cellcolor{SweepPurple!33.30}87.00
& 16 & \cellcolor{SweepPurple!34.05}87.83
& 16 & \cellcolor{SweepPurple!34.30}88.11
& 16 & \cellcolor{SweepPurple!31.60}85.11
& 16 & \cellcolor{SweepPurple!13.30}64.78
& 48 & \cellcolor{SweepPurple!34.25}88.06 \\

17 & \cellcolor{SweepPurple!35.70}89.67
& 17 & \cellcolor{SweepPurple!34.85}88.72
& 17 & \cellcolor{SweepPurple!34.80}88.67
& 17 & \cellcolor{SweepPurple!29.40}82.67
& 17 & \cellcolor{SweepPurple!16.70}68.56
& 49 & \cellcolor{SweepPurple!31.55}85.06 \\

18 & \cellcolor{SweepPurple!38.05}92.28
& 18 & \cellcolor{SweepPurple!38.00}\textbf{92.22}
& 18 & \cellcolor{SweepPurple!37.70}91.89
& 18 & \cellcolor{SweepPurple!27.25}80.28
& 18 & \cellcolor{SweepPurple!15.75}67.50
& 50 & \cellcolor{SweepPurple!33.45}87.17 \\

19 & \cellcolor{SweepPurple!38.25}92.50
& 19 & \cellcolor{SweepPurple!37.80}92.00
& 19 & \cellcolor{SweepPurple!37.95}\textbf{92.17}
& 19 & \cellcolor{SweepPurple!25.45}78.28
& 19 & \cellcolor{SweepPurple!17.55}69.50
& 51 & \cellcolor{SweepPurple!21.05}73.39 \\

20 & \cellcolor{SweepPurple!36.60}90.67
& 20 & \cellcolor{SweepPurple!37.45}91.61
& 20 & \cellcolor{SweepPurple!37.75}91.94
& 20 & \cellcolor{SweepPurple!0.00}45.61
& 20 & \cellcolor{SweepPurple!16.90}68.78
& 52 & \cellcolor{SweepPurple!2.65}52.94 \\

21 & \cellcolor{SweepPurple!38.05}92.28
& 21 & \cellcolor{SweepPurple!37.75}91.94
& 21 & \cellcolor{SweepPurple!37.25}91.39
& 21 & \cellcolor{SweepPurple!0.00}44.56
& 21 & \cellcolor{SweepPurple!16.55}68.39
& 53 & \cellcolor{SweepPurple!2.25}52.50 \\

22 & \cellcolor{SweepPurple!39.05}\textbf{93.39}
& 22 & \cellcolor{SweepPurple!36.65}90.72
& 22 & \cellcolor{SweepPurple!35.45}89.39
& 22 & \cellcolor{SweepPurple!0.00}45.67
& 22 & \cellcolor{SweepPurple!15.50}67.22
& 54 & \cellcolor{SweepPurple!0.00}48.56 \\

23 & \cellcolor{SweepPurple!24.30}77.00
& 23 & \cellcolor{SweepPurple!15.50}67.22
& 23 & \cellcolor{SweepPurple!13.00}64.44
& 23 & \cellcolor{SweepPurple!0.65}50.72
& 23 & \cellcolor{SweepPurple!17.25}69.17
& 55 & \cellcolor{SweepPurple!0.00}47.17 \\

24 & \cellcolor{SweepPurple!16.30}68.11
& 24 & \cellcolor{SweepPurple!10.05}61.17
& 24 & \cellcolor{SweepPurple!6.20}56.89
& 24 & \cellcolor{SweepPurple!1.45}51.61
& 24 & \cellcolor{SweepPurple!17.30}69.22
& 56 & \cellcolor{SweepPurple!0.00}47.67 \\

25 & \cellcolor{SweepPurple!18.70}70.78
& 25 & \cellcolor{SweepPurple!13.50}65.00
& 25 & \cellcolor{SweepPurple!10.70}61.89
& 25 & \cellcolor{SweepPurple!0.60}50.67
& 25 & \cellcolor{SweepPurple!16.40}68.22
& 57 & \cellcolor{SweepPurple!0.00}46.44 \\

26 & \cellcolor{SweepPurple!0.00}32.17
& 26 & \cellcolor{SweepPurple!0.00}35.06
& 26 & \cellcolor{SweepPurple!0.00}33.22
& 26 & \cellcolor{SweepPurple!0.00}49.83
& 26 & \cellcolor{SweepPurple!21.80}74.22
& 58 & \cellcolor{SweepPurple!0.00}47.72 \\

27 & \cellcolor{SweepPurple!0.00}40.50
& 27 & \cellcolor{SweepPurple!0.00}41.50
& 27 & \cellcolor{SweepPurple!0.00}39.83
& 27 & \cellcolor{SweepPurple!1.15}51.28
& 27 & \cellcolor{SweepPurple!33.05}86.72
& 59 & \cellcolor{SweepPurple!0.00}44.78 \\

28 & \cellcolor{SweepPurple!0.00}46.89
& 28 & \cellcolor{SweepPurple!2.20}52.44
& 28 & \cellcolor{SweepPurple!1.70}51.89
& 28 & \cellcolor{SweepPurple!0.00}49.50
& 28 & \cellcolor{SweepPurple!30.70}84.11
& 60 & \cellcolor{SweepPurple!0.00}45.56 \\

29 & \cellcolor{SweepPurple!0.00}48.83
& 29 & \cellcolor{SweepPurple!3.35}53.72
& 29 & \cellcolor{SweepPurple!4.15}54.61
& 29 & \cellcolor{SweepPurple!0.00}49.72
& 29 & \cellcolor{SweepPurple!28.05}81.17
& 61 & \cellcolor{SweepPurple!0.00}46.06 \\

30 & \cellcolor{SweepPurple!0.00}47.17
& 30 & \cellcolor{SweepPurple!1.55}51.72
& 30 & \cellcolor{SweepPurple!2.80}53.11
& 30 & \cellcolor{SweepPurple!0.45}50.50
& 30 & \cellcolor{SweepPurple!26.75}79.72
& 62 & \cellcolor{SweepPurple!0.00}44.89 \\

31 & \cellcolor{SweepPurple!0.00}48.44
& 31 & \cellcolor{SweepPurple!2.30}52.56
& 31 & \cellcolor{SweepPurple!3.35}53.72
& &
& 31 & \cellcolor{SweepPurple!35.55}89.50
& & \\

32 & \cellcolor{SweepPurple!1.05}51.17
& 32 & \cellcolor{SweepPurple!2.90}53.22
& 32 & \cellcolor{SweepPurple!3.20}53.56
& &
& &
& & \\

33 & \cellcolor{SweepPurple!3.60}54.00
& 33 & \cellcolor{SweepPurple!3.00}53.33
& 33 & \cellcolor{SweepPurple!4.85}55.39
& &
& &
& & \\

34 & \cellcolor{SweepPurple!4.45}54.94
& 34 & \cellcolor{SweepPurple!1.50}51.67
& 34 & \cellcolor{SweepPurple!0.25}50.28
& &
& &
& & \\

\bottomrule
\end{tabular}
\par
\endgroup
\caption{
\textbf{Layer sweeping accuracy on the COCO validation set.} Layer indices start at zero. Darker purple indicates higher accuracy, and bold marks the selected layer. Qwen3.6-27B is split into two consecutive layer ranges.
}
\label{tab:layer-sweep}
\end{table}

\FloatBarrier

\paragraph{Coordinates and correlations.}\label{app:coordinates}

For the correlations in Figure~\ref{fig:sspace-4}a and~\ref{fig:sspace-4}b, each object's horizontal and vertical ground truth coordinates come from the center of its annotated bounding box. For centering, we linearly map each normalized horizontal and vertical coordinate from \([0,1]\) to \([-1,1]\), and center the depth coordinate by subtracting its median. Each axis has 2,116 held-out COCO pairs. Absolute position correlations use the query and reference coordinates from every pair, giving 4,232 readouts. Relative position correlations use the 2,116 differences between query and reference coordinates.

\subsection{Reading out S-Space for Spatial QA}

We evaluate all five models at their selected layers using the same questions from each benchmark. For CV-Bench~\citep{tong2024cambrian}, we use all 1,250 questions in the ``Relation'' and ``Depth'' categories. Relation questions compare the horizontal or vertical positions of two objects. Depth questions ask which of two objects is closer to the camera.

For EmbSpatial~\citep{du2024embspatial}, we construct a subset of 1,200 questions from the evaluation data released by \citet{min2026whyfar}. We use a fixed random seed to select 200 questions for each of six directions (left, right, above, below, far, and close). For SpatialTunnel~\citep{min2026whyfar}, we use all 1,200 questions in the evaluation file, which contains 200 questions for each of the same six directions. Overall accuracy is the percentage of correctly answered questions in each benchmark.

\paragraph{Results.}
We compare direct generation and S-Space readouts for all five models on CV-Bench, EmbSpatial, and SpatialTunnel. Overall accuracy is reported in Table~\ref{tab:spatial-qa}, followed by results for each spatial direction in Tables~\ref{tab:cvbench-directions}, \ref{tab:embspatial-directions}, and~\ref{tab:spatialtunnel-directions}.

\begin{minipage}{\linewidth}
  \centering
  \small
  \setlength{\tabcolsep}{4pt}
  \begin{tabular*}{\linewidth}{@{\extracolsep{\fill}}llrrrrrr@{}}
    \toprule
    & & \multicolumn{2}{c}{\mirrosth{CV-Bench}} & \multicolumn{2}{c}{\mirrosth{EmbSpatial}} & \multicolumn{2}{c}{\mirrosth{SpatialTunnel}} \\
    \cmidrule(lr){3-4}\cmidrule(lr){5-6}\cmidrule(l){7-8}
    \mirrosth{Model} & \mirrosth{Layer} & \mirrosth{Direct} & \mirrosth{S-Space} & \mirrosth{Direct} & \mirrosth{S-Space} & \mirrosth{Direct} & \mirrosth{S-Space} \\
    \midrule
Molmo2-ER & 22 & \textbf{94.40} & 91.92 & \textbf{90.42} & 86.92 & \textbf{97.83} & 96.50 \\
MolmoAct2-Pretrain & 18 & \textbf{92.32} & 91.76 & 85.33 & \textbf{90.67} & 89.50 & \textbf{91.25} \\
MolmoAct2 & 19 & \textbf{91.76} & 91.44 & 88.58 & \textbf{89.92} & 89.00 & \textbf{92.75} \\
Qwen3.5-4B & 15 & 73.60 & \textbf{89.36} & 78.50 & \textbf{86.42} & 86.33 & \textbf{90.08} \\
Qwen3.6-27B & 43 & \textbf{93.76} & 91.84 & \textbf{91.33} & 89.00 & \textbf{96.83} & 92.92
 \\
    \bottomrule
  \end{tabular*}
  \captionof{table}{\textbf{Spatial QA accuracy (\%).} Direct generates answers, and S-Space compares object coordinates. Each model uses its layer selected on COCO. Bold marks the better method.}
  \label{tab:spatial-qa}
\end{minipage}

\par\medskip
\begin{minipage}{\linewidth}
\centering
\fontsize{9}{10.5}\selectfont
\renewcommand{\arraystretch}{1.08}
\setlength{\tabcolsep}{3.2pt}
\begin{tabular*}{\linewidth}{@{\extracolsep{\fill}}llrrrrrrr@{}}
\toprule
\mirrosth{Model} & \mirrosth{Method} & \mirrosth{Overall} & \mirrosth{Left} & \mirrosth{Right} & \mirrosth{Above} & \mirrosth{Below} & \mirrosth{Far} & \mirrosth{Close} \\
\midrule
Molmo2-ER & Direct & \textbf{94.40} & 89.07 & \textbf{97.00} & \textbf{94.44} & \textbf{95.93} & \textbf{96.33} & \textbf{93.33} \\
 & S-Space & 91.92 & \textbf{89.62} & 91.50 & \textbf{94.44} & \textbf{95.93} & 90.33 & 92.33 \\
\addlinespace[1.5pt]
MolmoAct2-Pretrain & Direct & \textbf{92.32} & \textbf{92.35} & \textbf{96.50} & \textbf{93.06} & \textbf{99.19} & \textbf{99.67} & 79.00 \\
 & S-Space & 91.76 & 88.52 & 92.00 & 90.97 & 94.31 & 92.67 & \textbf{92.00} \\
\addlinespace[1.5pt]
MolmoAct2 & Direct & \textbf{91.76} & \textbf{94.54} & \textbf{95.00} & \textbf{95.14} & \textbf{96.75} & \textbf{100.00} & 76.00 \\
 & S-Space & 91.44 & 90.71 & 90.00 & 90.28 & 91.06 & 91.33 & \textbf{93.67} \\
\addlinespace[1.5pt]
Qwen3.5-4B & Direct & 73.60 & \textbf{99.45} & 84.50 & \textbf{97.92} & 56.10 & \textbf{99.67} & 20.00 \\
 & S-Space & \textbf{89.36} & 89.62 & \textbf{89.00} & 94.44 & \textbf{81.30} & 85.67 & \textbf{94.00} \\
\addlinespace[1.5pt]
Qwen3.6-27B & Direct & \textbf{93.76} & \textbf{98.36} & \textbf{92.50} & \textbf{97.92} & 92.68 & \textbf{93.00} & 91.00 \\
 & S-Space & 91.84 & 90.16 & 91.50 & 95.14 & \textbf{95.93} & 84.33 & \textbf{97.33} \\
\bottomrule
\end{tabular*}
\captionof{table}{\textbf{CV-Bench accuracy splitted by answer direction.} S-Space uses the selected layer. Bold marks the better method within each model and column, including ties.}
\label{tab:cvbench-directions}
\end{minipage}

\par\medskip
\begin{minipage}{\linewidth}
\centering
\fontsize{9}{10.5}\selectfont
\renewcommand{\arraystretch}{1.08}
\setlength{\tabcolsep}{3.2pt}
\begin{tabular*}{\linewidth}{@{\extracolsep{\fill}}llrrrrrrr@{}}
\toprule
\mirrosth{Model} & \mirrosth{Method} & \mirrosth{Overall} & \mirrosth{Left} & \mirrosth{Right} & \mirrosth{Above} & \mirrosth{Below} & \mirrosth{Far} & \mirrosth{Close} \\
\midrule
Molmo2-ER & Direct & \textbf{90.42} & \textbf{92.50} & \textbf{95.50} & \textbf{97.50} & \textbf{97.50} & \textbf{88.50} & 71.00 \\
 & S-Space & 86.92 & \textbf{92.50} & 88.50 & \textbf{97.50} & 96.00 & 73.00 & \textbf{74.00} \\
\addlinespace[1.5pt]
MolmoAct2-Pretrain & Direct & 85.33 & 92.50 & \textbf{94.00} & 98.00 & \textbf{98.50} & \textbf{87.50} & 41.50 \\
 & S-Space & \textbf{90.67} & \textbf{95.50} & 92.50 & \textbf{98.50} & 97.00 & 79.50 & \textbf{81.00} \\
\addlinespace[1.5pt]
MolmoAct2 & Direct & 88.58 & \textbf{95.50} & \textbf{94.00} & 97.00 & \textbf{99.00} & \textbf{93.50} & 52.50 \\
 & S-Space & \textbf{89.92} & 92.50 & 90.00 & \textbf{98.50} & 96.50 & 78.00 & \textbf{84.00} \\
\addlinespace[1.5pt]
Qwen3.5-4B & Direct & 78.50 & \textbf{99.00} & 84.50 & \textbf{99.50} & 68.50 & \textbf{99.00} & 20.50 \\
 & S-Space & \textbf{86.42} & 90.50 & \textbf{94.00} & 98.50 & \textbf{98.00} & 47.50 & \textbf{90.00} \\
\addlinespace[1.5pt]
Qwen3.6-27B & Direct & \textbf{91.33} & \textbf{96.50} & 93.00 & 97.50 & 96.00 & \textbf{87.00} & 78.00 \\
 & S-Space & 89.00 & 93.50 & \textbf{93.50} & \textbf{99.00} & \textbf{97.50} & 59.50 & \textbf{91.00} \\
\bottomrule
\end{tabular*}
\captionof{table}{\textbf{EmbSpatial accuracy splitted by answer direction.} S-Space uses the selected layer. Bold marks the better method within each model and column, including ties.}
\label{tab:embspatial-directions}
\end{minipage}

\par\medskip
\begin{minipage}{\linewidth}
\centering
\fontsize{9}{10.5}\selectfont
\renewcommand{\arraystretch}{1.08}
\setlength{\tabcolsep}{3.2pt}
\begin{tabular*}{\linewidth}{@{\extracolsep{\fill}}llrrrrrrr@{}}
\toprule
\mirrosth{Model} & \mirrosth{Method} & \mirrosth{Overall} & \mirrosth{Left} & \mirrosth{Right} & \mirrosth{Above} & \mirrosth{Below} & \mirrosth{Far} & \mirrosth{Close} \\
\midrule
Molmo2-ER & Direct & \textbf{97.83} & 99.00 & \textbf{100.00} & \textbf{100.00} & 98.50 & \textbf{100.00} & 89.50 \\
 & S-Space & 96.50 & \textbf{99.50} & 99.00 & 99.50 & \textbf{100.00} & 85.50 & \textbf{95.50} \\
\addlinespace[1.5pt]
MolmoAct2-Pretrain & Direct & 89.50 & \textbf{100.00} & \textbf{100.00} & \textbf{100.00} & 99.00 & \textbf{100.00} & 38.00 \\
 & S-Space & \textbf{91.25} & 99.50 & 99.50 & 97.00 & \textbf{100.00} & 72.00 & \textbf{79.50} \\
\addlinespace[1.5pt]
MolmoAct2 & Direct & 89.00 & \textbf{100.00} & \textbf{100.00} & \textbf{100.00} & 99.00 & \textbf{100.00} & 35.00 \\
 & S-Space & \textbf{92.75} & 99.50 & 99.00 & 96.50 & \textbf{100.00} & 75.50 & \textbf{86.00} \\
\addlinespace[1.5pt]
Qwen3.5-4B & Direct & 86.33 & \textbf{100.00} & 95.50 & \textbf{100.00} & 94.50 & \textbf{100.00} & 28.00 \\
 & S-Space & \textbf{90.08} & \textbf{100.00} & \textbf{100.00} & \textbf{100.00} & \textbf{100.00} & 41.00 & \textbf{99.50} \\
\addlinespace[1.5pt]
Qwen3.6-27B & Direct & \textbf{96.83} & \textbf{100.00} & \textbf{100.00} & \textbf{100.00} & \textbf{100.00} & \textbf{94.50} & \textbf{86.50} \\
 & S-Space & 92.92 & \textbf{100.00} & 99.50 & \textbf{100.00} & \textbf{100.00} & 80.00 & 78.00 \\
\bottomrule
\end{tabular*}
\captionof{table}{\textbf{SpatialTunnel accuracy splitted by answer direction.} S-Space uses the selected layer. Bold marks the better method within each model and column, including ties.}
\label{tab:spatialtunnel-directions}
\end{minipage}

\par\medskip

\subsection{Causal Interventions}\label{section-intervention-controls}\label{app:writeback}\label{app:main-coordinate-interventions}

\paragraph{Editing hidden states.}
We edit an object's hidden state to change its S-Space coordinates. Let \(h\) denote this hidden state and \(V=[v_x,v_y,v_z]\) contain the spatial readout directions. To change the coordinates from \(\mathbf c=V^\top h\) to \(\mathbf c'\), we apply
\begin{equation}
B=V(V^\top V)^{-1},\qquad
h'=h+B(\mathbf c'-\mathbf c).
\end{equation}
This update gives \(V^\top h'=\mathbf c'\) when the three axes are linearly independent. It preserves the component of the hidden state orthogonal to S-Space.

\paragraph{Edits.}\label{app:norm-controls}
Let \(c_q\) and \(c_r\) denote the query and reference coordinates on the edited axis, and \(h_q,h_r\in\mathbb R^d\) their hidden states. We implement four edits:
\[
\begin{aligned}
\text{Shift one object:}\quad &c'_q=c_q+\beta,\qquad c'_r=c_r,\\
\text{Shift both objects:}\quad &c'_q=c_q+\beta,\qquad c'_r=c_r+\beta,\\
\text{Swap:}\quad &c'_q=c_q+\alpha(c_r-c_q),\quad c'_r=c_r-\alpha(c_r-c_q),\\
\text{Random edit:}\quad &h'_q=h_q+\alpha m_q u,\qquad h'_r=h_r-\alpha m_r u.
\end{aligned}
\]
Here, \(\beta\) is a signed offset. Shifting both objects preserves their coordinate difference \(c_q-c_r\). For swaps, \(\alpha=0\) leaves the coordinates unchanged, \(\alpha=1\) exchanges them, and larger values move them farther apart in the reversed order. We note that amplified intervention is a common practice in related literature of mechanistic analysis~\citep{gurnee2026verbalizable, kang2026linear}. The shifts and swaps keep the other two coordinates fixed.

For random edits, \(m_q,m_r\) are the norms of the query and reference swap updates at \(\alpha=1\). We multiply these norms by \(\alpha\) to set the random edit magnitudes:
\[
\begin{aligned}
&m_t=\lVert h_t^{\mathrm{swap}}(\alpha=1)-h_t\rVert_2,\quad t\in\{q,r\},\\
&u=\frac{g}{\lVert g\rVert_2},\quad g{\sim}\mathcal N(0,I).
\end{aligned}
\]
Here, \(h_t^{\mathrm{swap}}(\alpha=1)\) denotes the hidden state after a swap with \(\alpha=1\). The query and reference receive opposite directions, each with the same norm as the corresponding swap update at strength \(\alpha\). For each example, we keep \(u\) fixed across edit strengths.

\paragraph{Questions.}
We ask about absolute position, relative position, or object color using the following prompts. The position questions below showcase the horizontal axis.

\appendixprompt{1. Absolute position}{Is the \{query\} to the left or right of the image center? Answer with only one word.}
\appendixprompt{2. Relative position}{Is the \{query\} to the left or right of the \{reference\}? Answer with only one word.}
\appendixprompt{3. Object color}{What color is the \{query\}? Answer with only one word.}

\paragraph{Main results.}
We use Molmo2-ER for all interventions. Figure~\ref{fig:sspace-7} uses 800 EmbSpatial examples at L17. In Figure~\ref{fig:sspace-7}, swaps and random edits use \(\alpha=0,1,\ldots,6\), and shifts use \(\beta=0,\pm5,\ldots,\pm30\). Table~\ref{tab:causal-answer-retention} reports the answer retention values plotted in Figure~\ref{fig:sspace-7}.

\begin{minipage}{\linewidth}
\centering
\small
\setlength{\tabcolsep}{2pt}
\renewcommand{\arraystretch}{1.12}
\begin{tabular*}{\linewidth}{@{\extracolsep{\fill}}ll*{7}{c}@{}}
\toprule
\mirrosth{Edit method} & \mirrosth{Question} & \multicolumn{7}{c}{\mirrosth{Editing scale and answer retention (\%)}} \\
\midrule
\multirow{4}{*}{Shift one object}
& \multirow{2}{*}{Absolute position}
& \(\beta=0\) & \(\pm5\) & \(\pm10\) & \(\pm15\) & \(\pm20\) & \(\pm25\) & \(\pm30\) \\
& & 100.00 & 95.15 & 89.23 & 80.62 & 68.37 & 52.49 & 41.05 \\
\cmidrule(l){2-9}
& \multirow{2}{*}{Color}
& \(\beta=0\) & \(\pm5\) & \(\pm10\) & \(\pm15\) & \(\pm20\) & \(\pm25\) & \(\pm30\) \\
& & 100.00 & 97.00 & 93.50 & 89.88 & 86.38 & 82.50 & 78.13 \\
\midrule
\multirow{2}{*}{Shift both objects}
& \multirow{2}{*}{Relative position}
& \(\beta=0\) & \(\pm5\) & \(\pm10\) & \(\pm15\) & \(\pm20\) & \(\pm25\) & \(\pm30\) \\
& & 100.00 & 97.67 & 94.32 & 88.50 & 83.98 & 81.91 & 80.36 \\
\midrule
\multirow{2}{*}{Swap}
& \multirow{2}{*}{Relative position}
& \(\alpha=0\) & \(1\) & \(2\) & \(3\) & \(4\) & \(5\) & \(6\) \\
& & 100.00 & 98.71 & 85.66 & 59.82 & 51.16 & 45.87 & 43.93 \\
\midrule
\multirow{2}{*}{Random edit}
& \multirow{2}{*}{Relative position}
& \(\alpha=0\) & \(1\) & \(2\) & \(3\) & \(4\) & \(5\) & \(6\) \\
& & 100.00 & 100.00 & 99.91 & 99.91 & 99.87 & 99.83 & 99.83 \\
\bottomrule
\end{tabular*}
\captionof{table}{
\textbf{Answer retention under S-Space intervention} (Figure~\ref{fig:sspace-7}).
}
\label{tab:causal-answer-retention}
\end{minipage}

\par\medskip

\paragraph{Layer sweep.}\label{app:relative-controls}\label{app:color-controls}
The interventions at L17 show that editing S-Space changes spatial judgments. We repeat these interventions across layers to examine where this causal effect appears.

We use Molmo2-ER at L13--L20, with 80 examples per task at each layer. For Table~\ref{tab:relative-controls}, we apply the \emph{Swap} and \emph{Shift both objects} edits and ask \emph{Relative position} questions.
For Table~\ref{tab:absolute-color-controls}, we apply the \emph{Shift one object} edit along the horizontal axis and ask \emph{Absolute position} and \emph{Object color} questions.

\begin{minipage}{\linewidth}
\centering
\small
\setlength{\tabcolsep}{4pt}
\renewcommand{\arraystretch}{1.08}

\begin{tabular}{@{}lccc@{}}
\toprule
\mirrosth{Layer} & \mirrosth{Swap ($\alpha=5$)} & \mirrosth{Shift both object ($\beta=-30$)} & \mirrosth{Shift both object ($\beta=+30$)} \\
\midrule
13 & 73.75 & 85.00 & 81.25 \\
14 & 85.00 & 81.25 & 83.75 \\
15 & 50.00 & 75.00 & 80.00 \\
16 & 60.00 & 92.50 & 93.75 \\
17 & 62.50 & 93.75 & 93.75 \\
18 & 43.75 & 96.25 & 91.25 \\
19 & 85.00 & 96.25 & 95.00 \\
20 & 72.50 & 96.25 & 92.50 \\
\bottomrule
\end{tabular}
\captionof{table}{
\textbf{Layer-sweeping results for relative position interventions.}
Number refers to the accuracy (\%) of relative position questions under corresponding interventions on 80 examples per layer. Unedited accuracy is 97.50\%.
}
\label{tab:relative-controls}
\end{minipage}

\par\medskip
\begin{minipage}{\linewidth}
\centering
\small
\setlength{\tabcolsep}{10pt}
\renewcommand{\arraystretch}{1.08}

\begin{tabular}{@{}lcccc@{}}
\toprule
& \multicolumn{3}{c}{\mirrosth{Absolute position}} & \mirrosth{Color} \\
\cmidrule(lr){2-4}
\mirrosth{Layer}
& $|\beta|=10$
& $|\beta|=30$
& $|\beta|=50$
& $|\beta|=30$ \\
\midrule
13 & 1.25 & 30.00 & 76.25 & 11.25 \\
14 & 5.00 & 47.50 & 83.75 & 12.50 \\
15 & 11.25 & 75.00 & 93.75 & 17.50 \\
16 & 11.25 & 57.50 & 73.75 & 13.75 \\
17 & 10.00 & 52.50 & 55.00 & 11.25 \\
18 & 5.00 & 35.00 & 43.75 & 13.75 \\
19 & 2.50 & 8.75 & 6.25 & 8.75 \\
20 & 2.50 & 12.50 & 11.25 & 2.50 \\

\bottomrule
\end{tabular}
\captionof{table}{
\textbf{Layer-sweeping results for absolute position and color interventions.}
Number refers to the ratio (\%) of samples for which the answer changes under either a \(+\beta\) or a \(-\beta\) intervention at the corresponding layer.
}
\label{tab:absolute-color-controls}
\end{minipage}

\par\medskip

\subsection{Case Studies}\label{section-case-studies-appendix}

\begin{figure}[!htbp]
  \centering
  \includegraphics[width=0.80\linewidth]{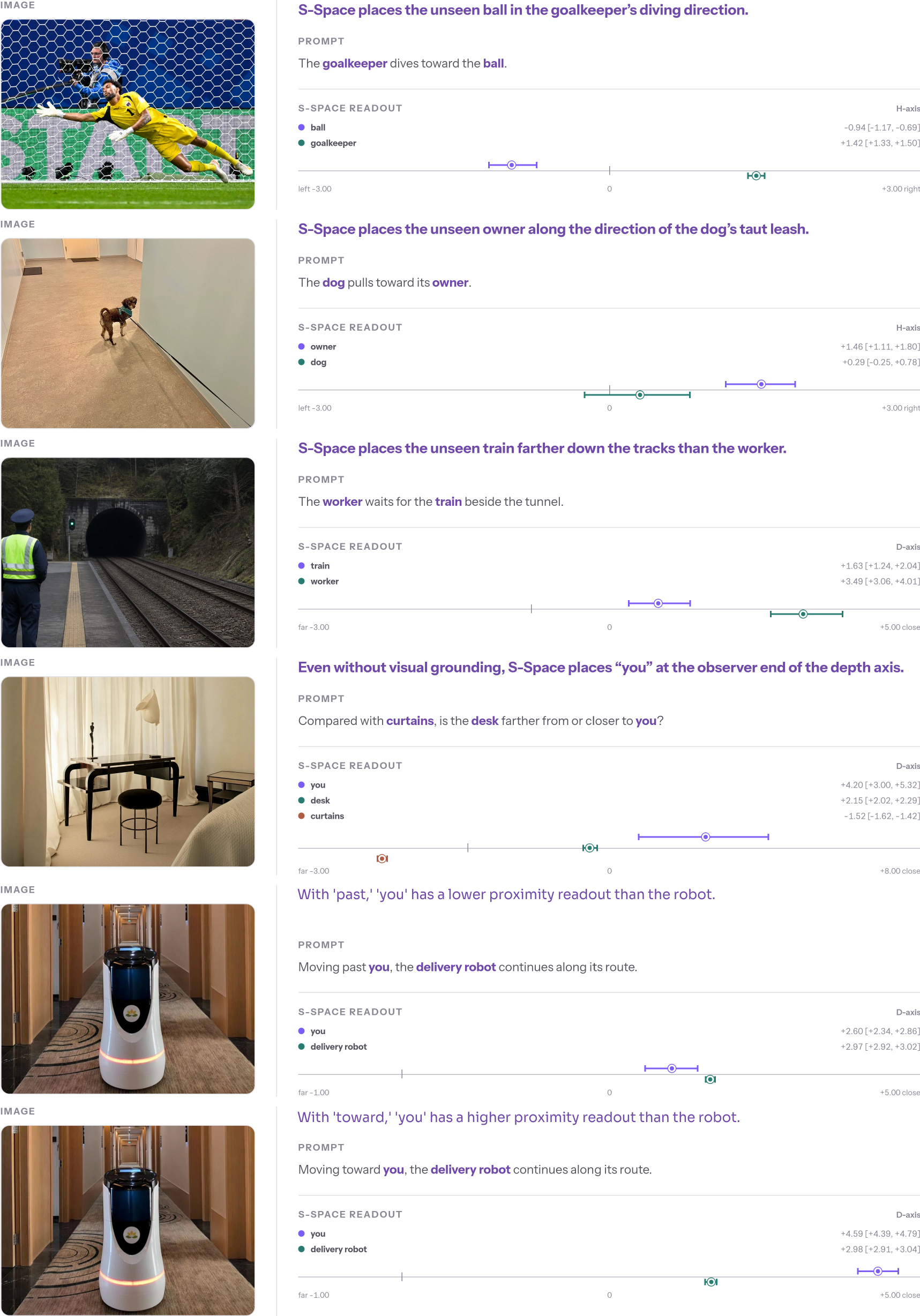}
  \caption{\textbf{S-Space beyond visible objects.} Images, prompts, and horizontal or distance readouts for an unseen ball, owner, train, the observer, and a robot moving past or toward. Readouts use Molmo2-ER L17. Means and 95\% bootstrap intervals are computed across prompt templates.}
  \label{fig:sspace-10}
\end{figure}

\paragraph{Unseen entities and the observer.}\label{section-beyond-visible-details}
We present additional examples of S-Space readouts for unseen entities and the observer in Figure~\ref{fig:sspace-10}.

\begin{figure}[!htbp]
  \centering
  \includegraphics[width=\linewidth]{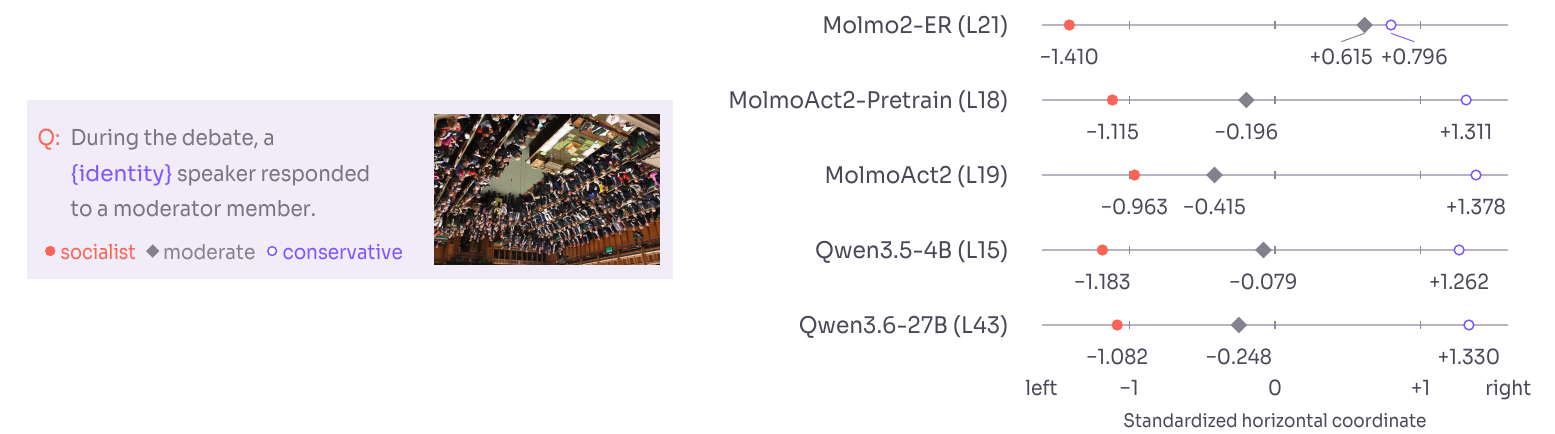}
  \caption{\textbf{Political semantics in S-Space.} Standardized horizontal readouts across five models.}
  \label{fig:political-all-models}
\end{figure}

\paragraph{Political semantics.}\label{section-political-details}
The political semantic results in Figure~\ref{fig:political-all-models} include MolmoAct2-Pretrain and MolmoAct2 alongside the three models in Figure~\ref{fig:case-language}, with the readouts standardized. All five models preserve the ordering of \emph{socialist}, \emph{moderate}, and \emph{conservative}.

\paragraph{Cardinal direction readouts.}\label{section-nsew-protocol}
We select 191 of the 1,000 MMSI-Bench questions~\citep{yang2026mmsi} by matching the correct option text to one of eight cardinal directions. The supplementary code includes the selected question IDs. We give Qwen3.6-27B the following instruction.

\appendixprompt{Cardinal direction readouts}{Give a concise step-by-step spatial explanation in at most five steps. End exactly with 'Answer: X', where X is the option letter.}

We replay each generated token sequence with the original images and prompt to extract hidden states at L43. For each occurrence of a direction word, we average the hidden states of its subtokens and project the result onto the horizontal and vertical axes identified on COCO. Compound directions written as one word, with a hyphen, or with a space are assigned to the same category and counted as a single occurrence.

Figure~\ref{fig:cardinal-diagnosis}b presents 1,311 occurrences from both correct and incorrect responses. For each direction, we first average its occurrences within each question, then average the results across questions with equal weight.

\FloatBarrier
\subsection{Perspective-Taking Spatial Reasoning}\label{app:reasoning}\label{section-spinbench-protocol}\label{app:computation}

\paragraph{Data and generation.}
We use the official SpinBench split consisting of 146 questions without a premise and 144 with a premise. Each question provides one image and asks which object would appear on the left or closer to the observer when viewed from the left, right, or back. Direct generation uses greedy decoding with a limit of 128 tokens. The Qwen Reasoning baselines use chain-of-thought generation with a limit of 32,768 tokens. S-Space + Rotation uses the layers selected on COCO in Table~\ref{tab:spatial-qa}. All methods use the same examples and object order.

\paragraph{Viewpoint computation.}
For objects \(A\) and \(B\) in the benchmark options, we compute the coordinate difference along axis \(g\) in the original view as
\[
m_g=v_g^\top(h_A-h_B).
\]
With \(x\) increasing to the right and \(z\) increasing toward the camera, we transform the coordinates using
\[
R_{\mathrm{left}}(x,z)=(z,-x),\qquad
R_{\mathrm{right}}(x,z)=(-z,x),\qquad
R_{\mathrm{back}}(x,z)=(-x,-z).
\]
After rotation, we select \(A\) if \(x'<0\) when the question asks which object appears on the left, or \(z'>0\) when it asks which object is closer to the observer. We select \(B\) otherwise. The implementation applies the following matrices.

\begin{appendixcode}
\ttfamily
\begin{verbatim}
VIEW_ROTATIONS = {
  "front": np.array([[ 1, 0,  0], [0, 1, 0], [ 0, 0,  1]]),
  "left" : np.array([[ 0, 0,  1], [0, 1, 0], [-1, 0,  0]]),
  "right": np.array([[ 0, 0, -1], [0, 1, 0], [ 1, 0,  0]]),
  "back" : np.array([[-1, 0,  0], [0, 1, 0], [ 0, 0, -1]]),
}

source_hvd = np.array([x, y, z])  # right, below, close
target_hvd = VIEW_ROTATIONS[target_view] @ source_hvd
\end{verbatim}
\end{appendixcode}

\paragraph{Controlled CoT.}\label{section-word-order-control}
For Figure~\ref{fig:sspace-15}, we supply the following reasoning sequence: the relation in the original view, the target viewpoint, and the relation after rotation. We fill in both relations using the benchmark labels.

\appendixprompt{Controlled CoT}{
The user wants me to determine \{TARGET\_QUESTION\} if the scene were viewed from the \{TARGET\_VIEW\}.\par\smallskip
1. **Analyze the current view:**\par
* In the front view, the \{QUERY\} is \{FRONT\_VIEW\_RELATION\} the \{REFERENCE\}.\par\smallskip
2. **Imagine the new perspective:**\par
* I now imagine viewing the \{QUERY\} and the \{REFERENCE\} from the \{TARGET\_VIEW\}.\par\smallskip
3. **Conclusion:**\par
* From this perspective, the \{QUERY\} is \{ROTATED\_VIEW\_RELATION\} the \{REFERENCE\}.
}

\paragraph{Main results.} SpinBench results for all five models are reported in Table~\ref{tab:spinbench-all-models}. The comparison covers Direct, Reasoning, and S-Space + Rotation, with and without spatial premises.
\par\noindent
\begin{minipage}{\linewidth}
  \centering
  \small
  \setlength{\tabcolsep}{4.0pt}
  \renewcommand{\arraystretch}{1.05}

  \begin{tabular*}{\linewidth}{@{\extracolsep{\fill}}lcccccc@{}}
    \toprule
    & \multicolumn{3}{c}{\mirrosth{Without premise ($n=146$)}}
    & \multicolumn{3}{c}{\mirrosth{With premise ($n=144$)}} \\
    \cmidrule(lr){2-4}\cmidrule(l){5-7}
    \mirrosth{Model}
    & \mirrosth{Direct}
    & \mirrosth{Reasoning}
    & \mirrosth{\shortstack{S-Space\\+ Rotation}}
    & \mirrosth{Direct}
    & \mirrosth{Reasoning}
    & \mirrosth{\shortstack{S-Space\\+ Rotation}} \\
    \midrule
    Molmo2-ER
& 60.96 & -- & \textbf{94.52}
& 61.81 & -- & \textbf{99.31} \\

MolmoAct2-Pretrain
& 66.44 & -- & \textbf{96.58}
& 56.25 & -- & \textbf{99.31} \\

MolmoAct2
& 71.23 & -- & \textbf{95.89}
& 64.58 & -- & \textbf{99.31} \\

Qwen3.5-4B
& 60.96 & 62.33 & \textbf{91.10}
& 60.42 & 72.22 & \textbf{97.92} \\

Qwen3.6-27B
& 60.96 & 83.56 & \textbf{95.21}
& 72.92 & 91.67 & \textbf{99.31} \\
    \bottomrule
  \end{tabular*}
  \captionof{table}{\textbf{SpinBench accuracy across all five models.} Direct generates an answer without intermediate reasoning. Reasoning uses chain-of-thought. S-Space + Rotation applies the requested viewpoint rotation to the readout coordinates. A premise supplies the correct spatial relation in the original view. Dashes indicate no reasoning mode. Bold marks the best method.}
  \label{tab:spinbench-all-models}
\end{minipage}

\par\medskip

\paragraph{S-Space with additional evidence.}
For Molmo2-ER, we further compare S-Space accuracy across layers with and without spatial premises. As shown in Figure~\ref{fig:sspace-16}, with the premise included, the coordinates read from S-Space become more accurate, allowing the readout-and-rotation pipeline to answer perspective-taking questions more reliably.

\par\noindent
\begin{minipage}{\linewidth}
  \centering
  \begin{minipage}[t]{0.2306\linewidth}\vspace{0pt}\centering \includegraphics[width=\linewidth]{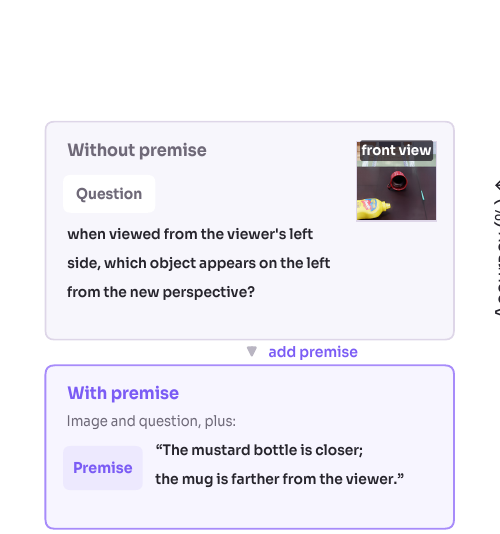}\par\smallskip {\small\textbf{(a)} Source-view premise}\end{minipage}\begin{minipage}[t]{0.3555\linewidth}\vspace{0pt}\centering \includegraphics[width=\linewidth]{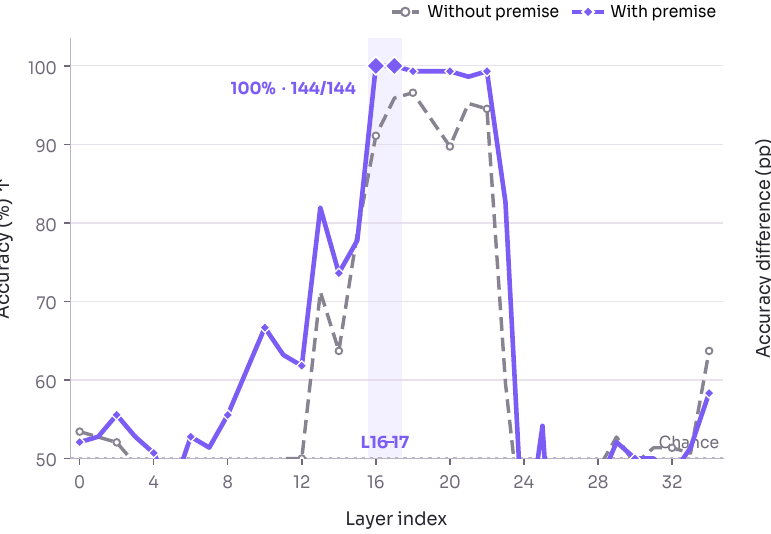}\par\smallskip {\small\textbf{(b)} Layerwise accuracy}\end{minipage}\begin{minipage}[t]{0.2239\linewidth}\vspace{0pt}\centering \includegraphics[width=\linewidth]{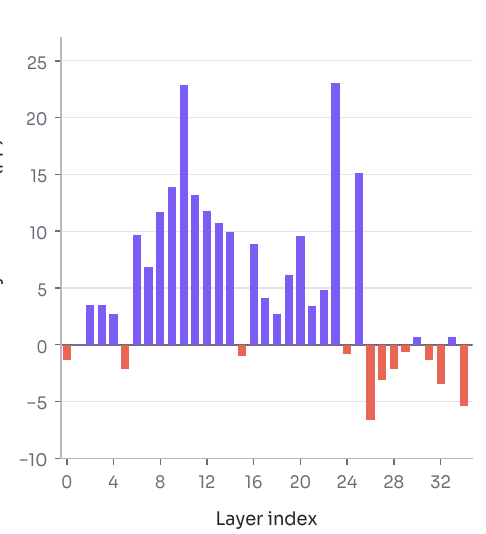}\par\smallskip {\small\textbf{(c)} Accuracy difference}\end{minipage}
  \captionof{figure}{\textbf{Molmo2-ER readouts with spatial premises.} \textbf{(a)} Source-view premise. \textbf{(b)} Layerwise accuracy with ($n=144$) and without ($n=146$) premises. \textbf{(c)} Accuracy difference.}
  \label{fig:sspace-16}
\end{minipage}
\par\medskip

\subsection{S-Space under Action Training}\label{section-additional-probes}\label{section-action-supervision-sharpens-interaction-geometry}\label{section-how-does-s-space-emerge}

As shown in Section~\ref{section-case-studies}, language translates space into words and shapes the emergence of S-Space. Action, by contrast, translates space into movement, providing a more precise interface with the physical world. To reach, grasp, or manipulate an object, a model must resolve both the geometry of the scene and the functional parts through which interaction occurs. We therefore ask whether learning to act sharpens the geometry in S-Space by comparing a VLM with its fine-tuned VLA models.

\paragraph{Models and data.} We examine Molmo2-ER, MolmoAct2-Pretrain, and MolmoAct2 on images from the InstructPart dataset~\citep{wan2025instructpart}, in which a manipulable object contains a characteristic functional part with estimated depth of the part and the whole object. We construct prompts such as ``... Locate the [object]" and assess the S-Space readout on the distance axis by comparison between the part and the whole, as in Section~\ref{sec:method-readout}. Ground-truth depth comparison use the provided object and part masks and Depth Anything V2-Small~\cite{yang2024depthanythingv2}.

\paragraph{Results.} As shown in Figure~\ref{fig:sspace-13}, we find an orderly progression along the model family: from the VLM through the action-model pretrain to the VLA, this local geometry becomes increasingly faithful to the physical relation between them.

\par\medskip\noindent
\begin{minipage}{\linewidth}
\centering
  \begin{minipage}[t]{0.6552\linewidth}\vspace{0pt}\centering\includegraphics[width=\linewidth]{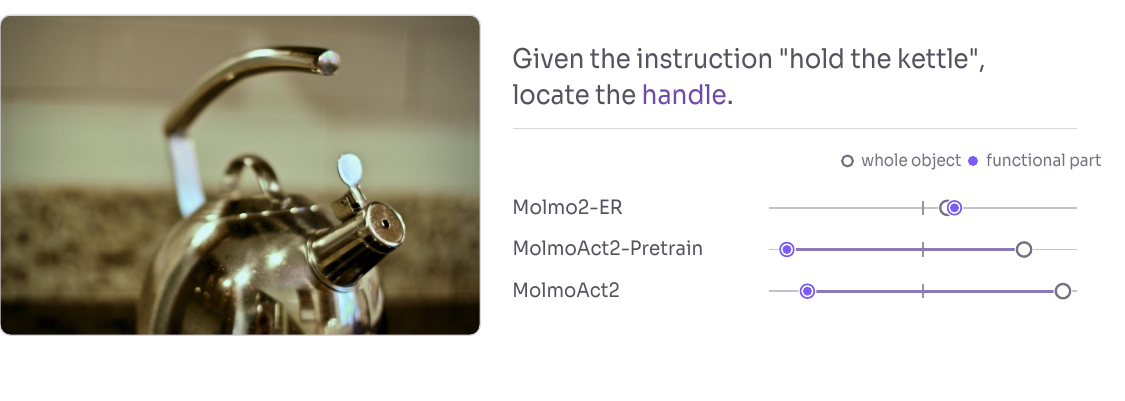}\par\smallskip{\small\textbf{(a)} Object and part readouts}\end{minipage}\begin{minipage}[t]{0.3448\linewidth}\vspace{0pt}\centering\includegraphics[width=\linewidth]{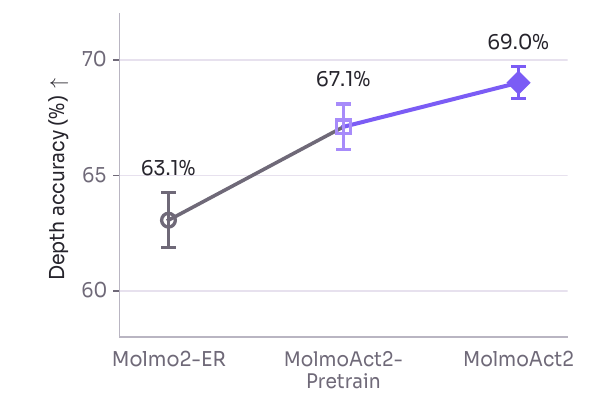}\par\smallskip{\small\textbf{(b)} Depth-comparison accuracy}\end{minipage}
  \par\smallskip
  \captionof{figure}{\textbf{Action training sharpens part–whole depth geometry in S-Space.} \textit{(a)} Depth readouts for the kettle and its handle. \textit{(b)} Depth comparison accuracy at L18 for all three models. Error bars indicate 95\% confidence intervals across prompts.}
  \label{fig:sspace-13}
\end{minipage}

\subsection{S-Space in Multi-View Contexts}\label{app:image-order}

Although S-Space is identified entirely from single-view examples, a workspace should still provide meaningful spatial coordinates when multiple views of the same scene are presented. We examine whether the model can use the prompt to select which view defines the observer-centered frame.

\paragraph{View pairs and prompt steering.}
We construct 600 view pairs in panoramic scenes from~\citet{yu2025thinking360deghumanoidvisual}, with 100 per direction. The same object appears in both views. As illustrated in the prompt-steering test (Figure~\ref{fig:hos-multiview-showcases}a), we keep the prompt fixed on ``In View 1'' while swapping which image is designated as View 1, and test whether the resulting change in an object's S-Space coordinates follows the corresponding change in viewpoint.

\paragraph{Results.} Across models, steering accuracy is consistently well above random choice along the horizontal, vertical, and distance axes (Figure~\ref{fig:hos-multiview-showcases}b). These results suggest that S-Space is not tied to a single visual input, but can be dynamically anchored to the viewpoint selected through language.

\par\medskip\noindent
\begin{minipage}{\linewidth}
\centering \includegraphics[width=\linewidth]{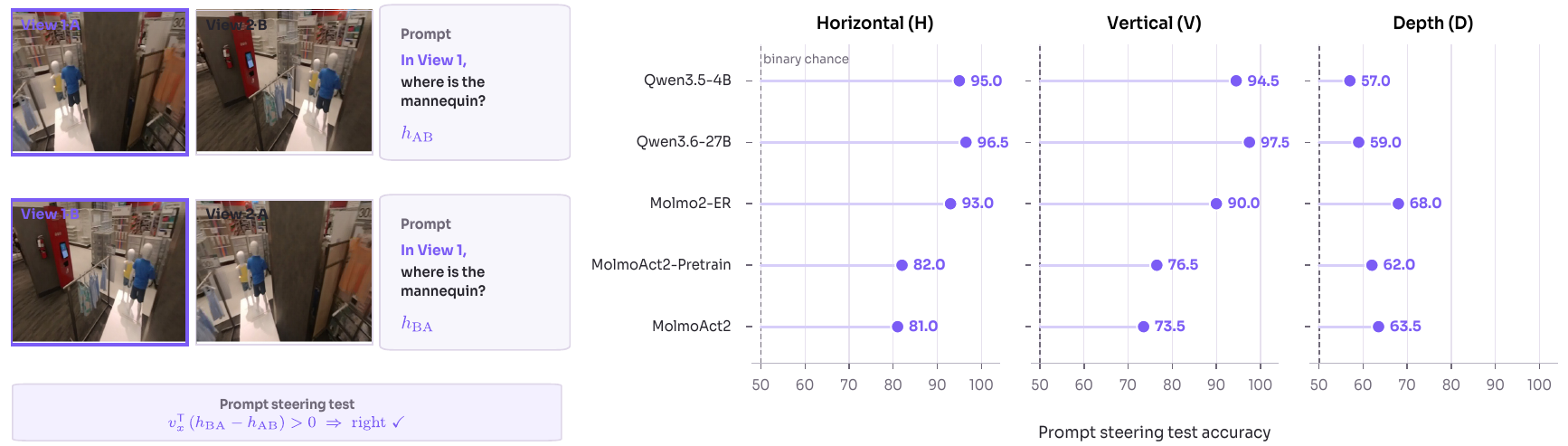} \par\smallskip\noindent \makebox[0.475\linewidth][c]{\small\textbf{(a)} Prompt steering test}\makebox[0.525\linewidth][c]{\small\textbf{(b)} Steer accuracy} \captionof{figure}{\textbf{Prompt steering test of S-Space in multi-view contexts.} \textit{(a)} Each example is evaluated twice with the same View-1 query while the two input images are swapped. \textit{(b)}
We evaluate the sign agreement between the resulting change in the S-Space coordinates and the object's ground-truth displacement between the two views.} \label{fig:hos-multiview-showcases}
\end{minipage}

\end{document}